\documentclass[acmtog,nonacm]{acmart}
\AtBeginDocument{%
  }

\usepackage{float}
\usepackage{graphicx}
\usepackage{wrapfig}
\usepackage{multirow}
\usepackage{tabularx}
\usepackage{makecell}
\usepackage{booktabs}        
\usepackage{caption}
\usepackage{subcaption}
\usepackage{colortbl}
\usepackage{utfsym}
\usepackage{listings}
\usepackage[most]{tcolorbox}
\usepackage{multicol}
\usepackage[normalem]{ulem}

\definecolor{best}{rgb}{1.0, 0.85, 0.6}      
\definecolor{second}{rgb}{0.7, 0.9, 1.0}     
\definecolor{sh_blue}{rgb}{0,0.60,0.93}
\definecolor{sh_gray2}{rgb}{1,0.89,0.75}
\definecolor{lyellow}{rgb}{1,0.63,0.098}
\definecolor{lred}{rgb}{0.906,0.42,0.32}
\definecolor{color3}{rgb}{0.95,0.95,0.95}
\definecolor{mygray}{gray}{.9}
\definecolor{genhaze}{rgb}{0.60, 0.57, 0.79}
\definecolor{bluegreen}{rgb}{0.44, 0.64, 0.77}
\definecolor{gray_venue}{rgb}{0.53,0.52,0.52}
\definecolor{color5}{rgb}{1,0.96,0.88}

\newlength{\Oldarrayrulewidth}

\lstdefinelanguage{json}{
  basicstyle=\ttfamily\small,
  breaklines=true,
  morecomment=[l]{//},
  morestring=[b]",
  stringstyle=\color{red!70!black},
  commentstyle=\color{gray!70!white},
  literate=
    *{0}{{{\color{gray}{0}}}}1
     {1}{{{\color{gray}{1}}}}1
     {:}{{{\color{black}{:}}}}1
     {,}{{{\color{black}{,}}}}1
     {\{}{{{\color{blue}{\{}}}}1
     {\}}{{{\color{blue}{\}}}}}1
     {[}{{{\color{blue}{[}}}}1
     {]}{{{\color{blue}{]}}}}1
}

\tcbuselibrary{listingsutf8} 
\tcbset{
    examplebox/.style={
        colback=green!5!white, 
        colframe=green!75!black, 
        coltitle=green!50!black, 
        fonttitle=\bfseries, 
        boxrule=0.5mm, 
        sharp corners, 
        enhanced, 
        attach boxed title to top left={
            yshift=-2mm, 
            xshift=5mm 
        },
        boxed title style={
            colframe=green!75!black, 
            colback=white, 
            sharp corners 
        }
    }
}

\newtcolorbox[auto counter, number within=section]{example}[2][]{%
    examplebox,
    title=Prompt   ~\thetcbcounter~(#2), 
    #1 
}
\begin{document}

\title{PosterCraft: Rethinking High-Quality Aesthetic Poster Generation in a Unified Framework}


\author{Sixiang Chen$^{1,2,*}$, Jianyu Lai$^{1,*}$, Jialin Gao$^{2,*}$, Tian Ye$^{1}$, Haoyu Chen$^{1}$, Hengyu Shi$^{2}$, Shitong Shao$^{1}$, Yunlong Lin$^{3}$, Song Fei$^{1}$, Zhaohu Xing$^{1}$, Yeying Jin$^{4}$, Junfeng Luo$^{2}$, Xiaoming Wei$^{2}$, Lei Zhu$^{1,5,\dagger}$}

\affiliation{%
  \institution{
  \\$^{1}$ The Hong Kong University of Science and Technology (Guangzhou), $^{2}$ Meituan, $^{3}$ Xiamen University, $^{4}$ National University of Singapore, $^{5}$ The Hong Kong University of Science and Technology \\$^{*}$ Equal Contribution; $^{\dagger}$ Corresponding Author}
  }

\begin{abstract}
Generating aesthetic posters is more challenging than simple design images: it requires not only precise text rendering but also the seamless integration of abstract artistic content, striking layouts, and overall stylistic harmony. To address this, we propose PosterCraft, a unified framework that abandons prior modular pipelines and rigid, predefined layouts, allowing the model to freely explore coherent, visually compelling compositions. PosterCraft employs a carefully designed, cascaded workflow to optimize the generation of high-aesthetic posters: (i) large-scale text-rendering optimization on our newly introduced Text-Render-2M dataset; (ii) region-aware supervised fine-tuning on HQ-Poster-100K; (iii) aesthetic-text reinforcement learning via best-of-n preference optimization; and (iv) joint vision–language feedback refinement. Each stage is supported by a fully automated data-construction pipeline tailored to its specific needs, enabling robust training without complex architectural modifications. Evaluated on multiple experiments, PosterCraft significantly outperforms open-source baselines in rendering accuracy, layout coherence, and overall visual appeal—approaching the quality of SOTA commercial systems. Our code, models, and datasets can be found in the {\tt Project page: \url{https://ephemeral182.github.io/PosterCraft/}}

\end{abstract}

\begin{teaserfigure}
\setlength{\abovecaptionskip}{0.2cm} 
\setlength{\belowcaptionskip}{-0.0cm}
  \includegraphics[width=\textwidth]{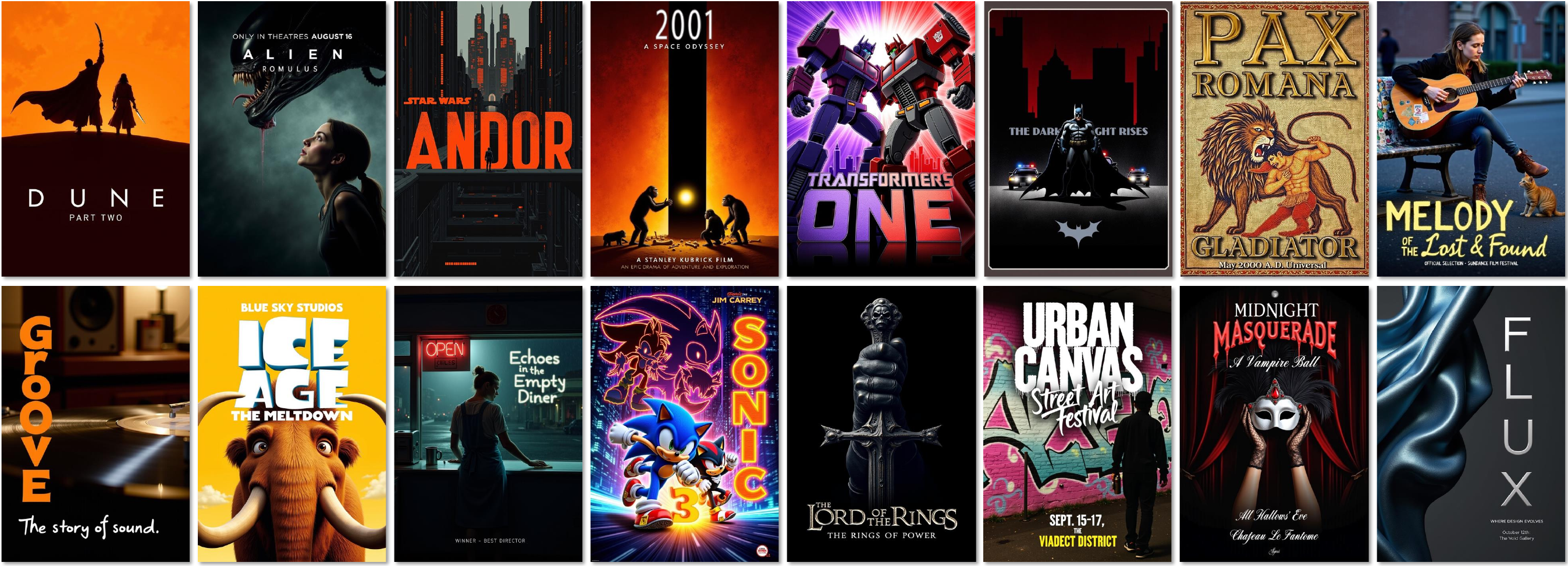}
  \caption{Aesthetic posters generated by PosterCraft demonstrate that backgrounds, layouts, and typographic designs are produced directly from textual input without modular designs, highlighting its ability to employ a unified framework to generate posters with visual consistency and compelling aesthetic appeal.}
  \Description{Enjoying the baseball game from the third-base
  seats. Ichiro Suzuki preparing to bat.}
  \label{fig:teaser}
\end{teaserfigure}

\received{20 February 2007}
\received[revised]{12 March 2009}
\received[accepted]{5 June 2009}

\maketitle

\vspace{-0.2cm}
\section{Introduction}
\label{headings}
Despite recent advances in automated visual design, aesthetic 
poster generation remains a formidable challenge and is still relatively 
underexplored.
Existing generative approaches primarily focus on foundational tasks such as 
text rendering~\cite{textdiffuser,glyph,anytext,chen2023textdiffuser} or the creation of specific product-oriented posters~\cite{wang2025designdiffusion,gao2025postermaker,podell2023sdxl}, offering 
limited capacity to produce high-quality, aesthetically compelling outputs.
These methods often fall short of addressing the multifaceted demands of 
aesthetic poster design, which requires not only \textit{\textbf{(i)}} accurate and 
stylistically coherent text, but also \textit{\textbf{(ii)}} the creation of 
abstract and visually appealing artistic content and \textit{\textbf{(iii)}} the striking 
layouts and holistic stylistic consistency.
Therefore, aesthetic poster generation demands a more comprehensive synthesis 
of content, form, and communicative intent.

Recent approaches to aesthetic poster generation~\cite{chen2025posta,yang2024posterllava,seol2024posterllama,tang2023layoutnuwa} have primarily followed a modular design paradigm. 
Typically, a fine-tuned vision-language model (VLM) acts as a layout planner, suggesting text content and positioning. 
The suggestions are then overlaid onto a separately generated background, or used as hard constraints for the generative model to follow.
However, this design strategy presents several limitations. 
\textit{\textbf{(i.) Lack of aesthetic consistency:}} it undermines the visual and stylistic coherence essential for aesthetic poster creation. 
\textit{\textbf{(ii.) Limited visual quality:}} it constrains the upper bound of visual quality due to the decoupled design process and heavy reliance on the VLM's accuracy and robustness.
In contrast, existing end-to-end design-centered generation approaches~\cite{textdiffuser,inoue2024opencole,wang2025designdiffusion,gao2025postermaker} 
remain limited to relatively simple tasks, 
such as greeting cards or product compositions, which lack the visual and structural complexity of high-quality aesthetic posters. Additionally, while powerful foundation models~\cite{flux,sd3,stabilityai2024sd35large} have demonstrated impressive capabilities in generating complex natural images, they still fall short of meeting all the specific requirements of aesthetic posters. (e.g. precise text rendering, abstract artistic content, and holistic stylistic coherence).
For this, we classify it as \textit{\textbf{(iii.) Simplified use cases}}.
More importantly, the absence of large-scale, versatile datasets tailored specifically for aesthetic poster generation has further constrained 
the development of fully generative solutions---\textit{\textbf{(iv.) Absence of targeted datasets}}.

To move beyond the limitations of current modular and simply scoped generative paradigms, we leverage the capabilities of foundation models to explore unified generation for aesthetic posters, aiming to produce visually coherent and artistically compelling results.
In this work, we argue that incremental, 
component-level improvements alone are insufficient to achieve significant aesthetic gains. 
Instead, we propose a unified framework, \textbf{\textit{PosterCraft}}, which includes a comprehensive workflow to systematically perform four critical stages:
\textit{\textbf{(i.)}} scalable text rendering optimization, \textit{\textbf{(ii.)}} high-quality poster fine-tuning, \textit{\textbf{(iii.)}} aesthetic-text reinforcement learning, and \textit{\textbf{(iv.)}} vision-language feedback refinement. 
To support this workflow, we construct a suite of specialized datasets for each stage through automated pipelines, 
enabling robust training and facilitating future research in aesthetic poster generation. 
This framework empowers the trained model to generate high-quality posters in the end-to-end pass. 
Experiments demonstrate that our approach significantly outperforms existing baselines and achieves competitive 
results compared to several closed-source models.

Overall, our contributions can be summarized as follows:
\vspace{-0.2cm}
\begin{itemize}
    \item \textbf{A unified framework for aesthetic poster generation:} We revisit aesthetic poster generation through an end-to-end approach tailored for high-quality, visually coherent posters, surpassing prior modular pipelines and methods focused on simpler or product-centric designs.
    \item \textbf{A cascade workflow for high-quality poster optimization:} We propose a unified training pipeline with four stages: (i) scalable text rendering optimization, (ii) high-quality poster fine-tuning, (iii) aesthetic-text reinforcement learning, and (iv) vision-language feedback refinement. Each stage targets a key challenge in aesthetic poster generation, enabling the model to produce artistically compelling results at inference time.
    \item \textbf{Stage-specific, fully automated dataset construction:} We construct specialized datasets for each workflow stage using automated collection and filtering, tailored to the unique demands of aesthetic poster generation. These datasets overcome the limitations of resources and support more robust, transferable training.
    \item \textbf{Superior performance over existing baselines:} Extensive experiments show that our method significantly 
    outperforms open-source baselines in terms of both aesthetic quality and layout structure, and achieves competitive 
    performance compared to commercial systems.
\end{itemize}

\begin{figure*}[t] 
\centering
\setlength{\abovecaptionskip}{0.1cm} 
\setlength{\belowcaptionskip}{-0.1cm}
\includegraphics[width=1\textwidth]{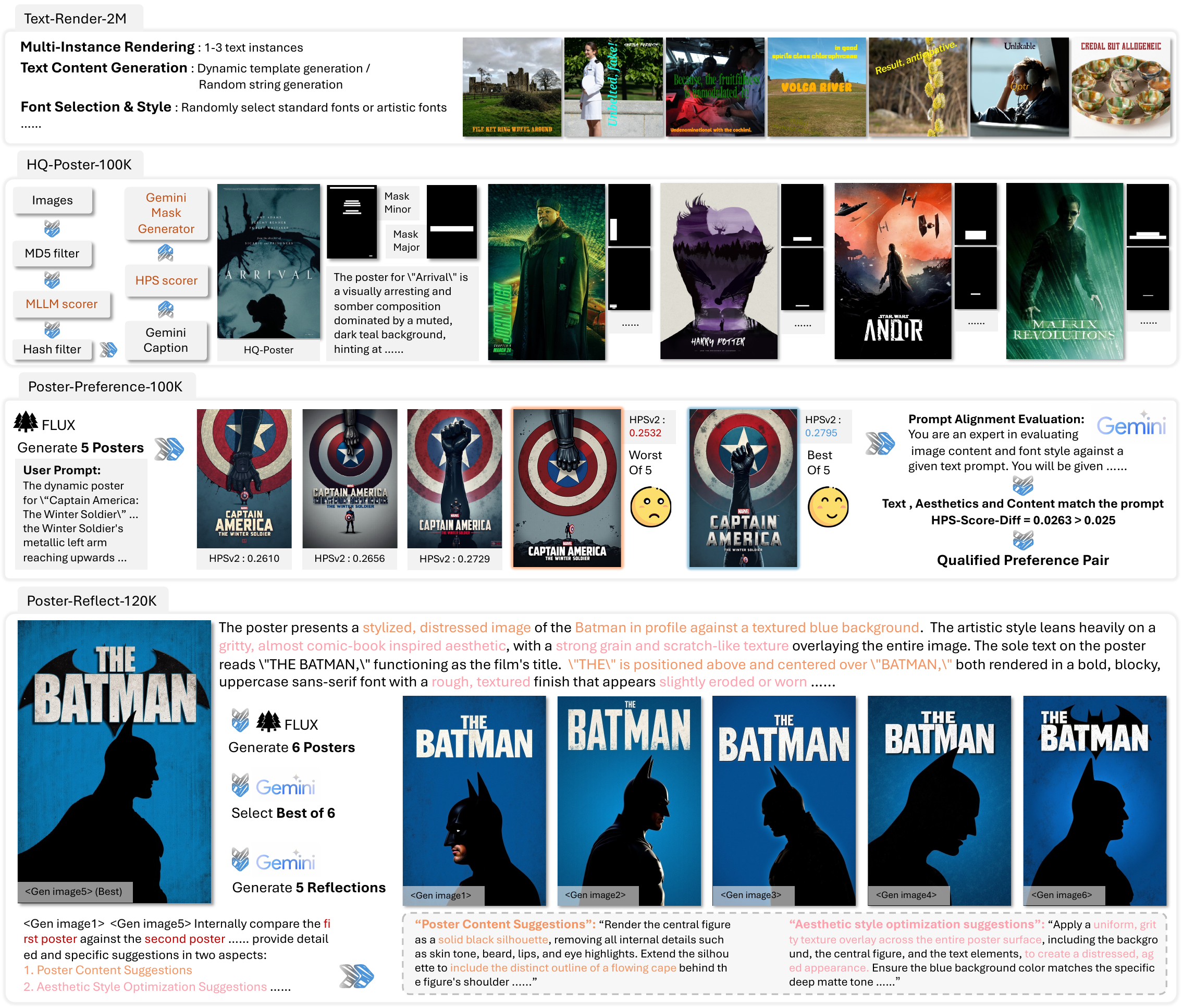}
\caption{\textbf{The four datasets of PosterCraft} across its four stages: (1) Text-Render-2M for text rendering optimization in the initial phase, (2) HQ-Poster-100K, comprising over 100K high-quality posters with masks and captions, (3) Poster-Preference-100K, yielding 6K high-quality preference pairs from 100K generated samples, and (4) Poster-Reflect-120K, constructing 64K feedback pairs from 120K generated posters.}
\label{dataset fig}
\end{figure*}

\begin{figure*}[t] 
\centering
\setlength{\abovecaptionskip}{0.1cm} 
\setlength{\belowcaptionskip}{-0.1cm}
\includegraphics[width=1\textwidth]{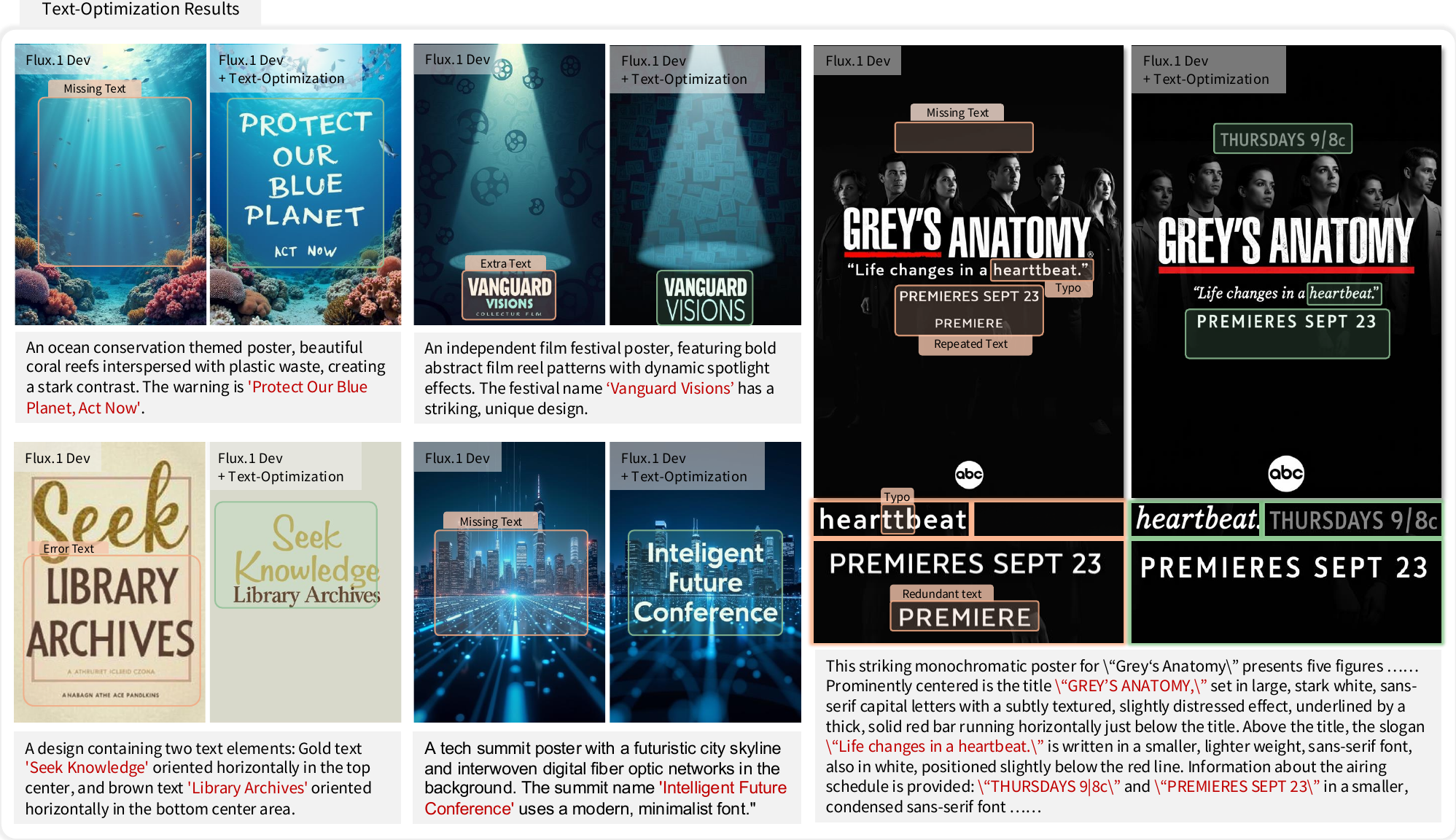}
\caption{\textbf{Comparison of text rendering on poster typography, plain-text scenes, and long-form text posters}. Each pair shows the Flux.1 dev baseline (left), exhibiting missing, repeated, or error text, alongside the optimized output (right) after our scalable text rendering optimization, demonstrating marked gains in text fidelity, alignment, and accuracy.}
\label{text rendering effect}
\end{figure*}

\section{Related Works}

\noindent{\textbf{Image Generation for Design Images.}}
Recently, design-centric image generation has transitioned from early GAN- and VAE-based generators to more powerful diffusion-based frameworks~\cite{textdiffuser,anytext,wang2025designdiffusion,gao2025postermaker,inoue2023layoutdm,zheng2023layoutdiffusion,chen2023textdiffuser}. Notably, LayoutDiffusion~\cite{zheng2023layoutdiffusion} reformulated layout generation as a discrete token denoising process, achieving significant gains. 
Advanced pipelines like TextDiffuser~\cite{textdiffuser,chen2023textdiffuser} adopted a two-stage approach: a transformer-based layout planner followed by a diffusion model conditioned on OCR-derived masks to generate coherent text images from prompts.
DesignDiffusion~\cite{wang2025designdiffusion} improved text-centric design by enhancing prompts with character-level embeddings and applying a localization loss to boost text rendering accuracy.
Despite recent progress, many methods imposed rigid pre-layout constraints on text and layout, often undermining overall aesthetic coherence. Furthermore,  their focus on specific domains such as product ads or greeting cards also limited the complexity and creativity of the tasks they address. In contrast, our approach explores a unified framework for aesthetic poster generation, integrating text rendering, artistic content creation, and visually striking layout design within a single inference process.

\noindent{\textbf{VLM for Image Generation.}}
Vision–language models served dual roles as both high-level planners ~\cite{lin2023layoutprompter,luo2024layoutllm,feng2023layoutgpt,yang2024posterllava,rpg,chen2025posta,tang2023layoutnuwa} and fully end-to-end generators ~\cite{zhou2024transfusion,xie2024show,team2024chameleon,ma2024janusflow,zhou2024transfusion,wu2024janus}.
LayoutGPT ~\cite{feng2023layoutgpt} translated natural-language instructions into structured layout specifications, while fine-tuned adaptations PosterLlama ~\cite{seol2024posterllama} and PosterLLava ~\cite{yang2024posterllava} employed HTML- or JSON-formatted tokens to produce content-aware graphic designs.
POSTA ~\cite{chen2025posta} further leveraged a fine-tuned VLM as an "aesthetic designer", applying modular design atop existing high-quality images.
Unified transformer architectures like TransFusion ~\cite{zhou2024transfusion}, and JanusFlow ~\cite{ma2024janusflow} advanced this paradigm by generating image and text tokens in one architecture, enabling one-shot synthesis of complete visual compositions.
Complementing these approaches, feedback-driven pipelines ~\cite{li2024enhancing,li2025reflect} incorporated VLM-based critics to assess object accuracy during training.
Nevertheless, these VLM-driven methods continued to struggle with maintaining cohesive aesthetic consistency, and faced an upper bound in layout complexity imposed by the models' architectural capacity and the scope of their limited training data.

\begin{figure*}[t] 
\centering
\setlength{\abovecaptionskip}{0.2cm} 
\setlength{\belowcaptionskip}{-0.3cm}
\includegraphics[width=1\textwidth]{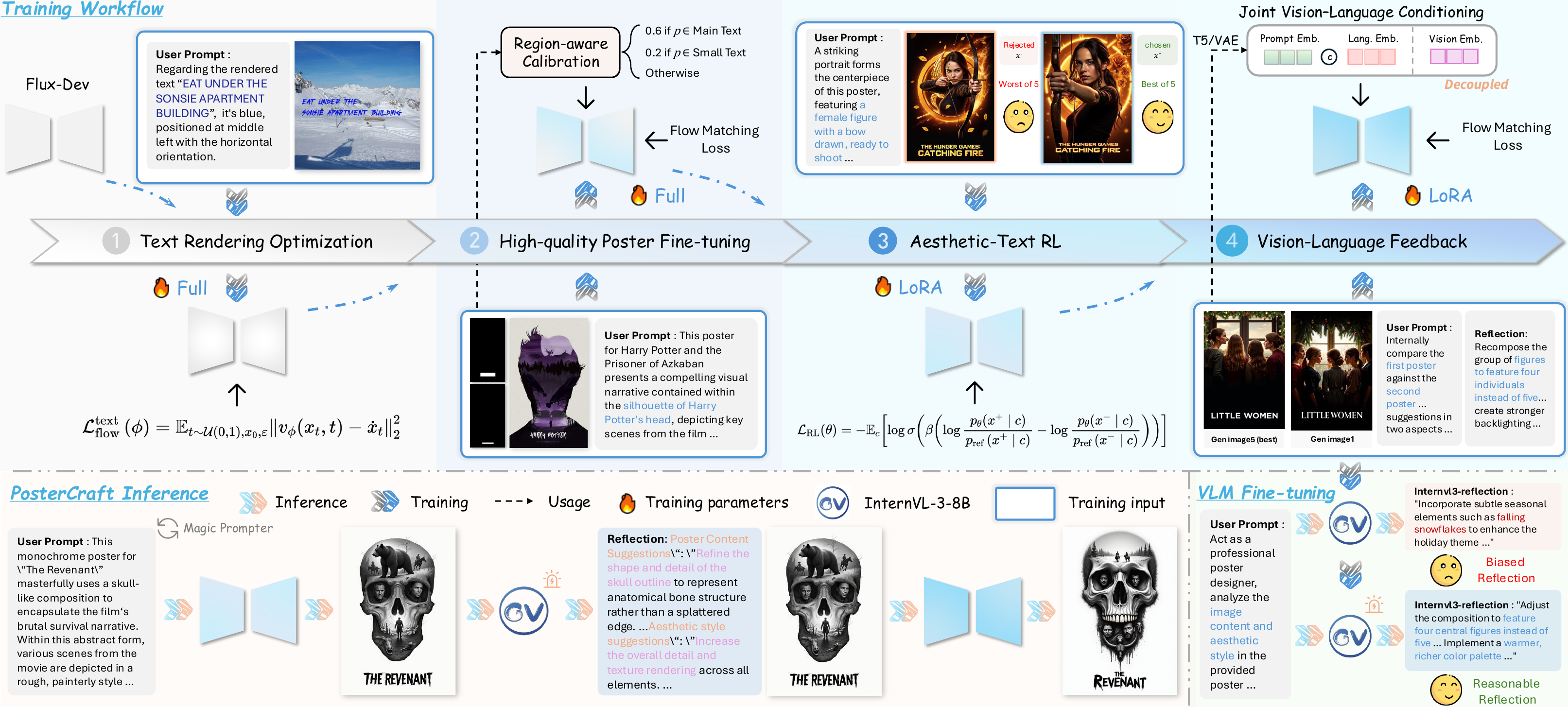}
\caption{\textbf{The pipeline of PosterCraft}, which has four stages: (1) Text Rendering Optimization to improve text accuracy and fidelity; (2) High-Quality Poster Fine-Tuning with region-aware calibration to poster styling across text and non-text regions; (3) Aesthetic-Text Reinforcement Learning to instill detailed aesthetic and content preferences; and (4) Joint Vision–Language Feedback, integrating multimodal reflections for refined outputs. At inference, the fine-tuned model generates high-quality aesthetic posters end-to-end from a single prompt, with an optional VLM-driven critique loop.}

\label{overview}
\end{figure*}

\vspace{-0.2cm}
\section{Unified Workflow and Specific Dataset}
In this work, we rethink aesthetic poster generation: a high-capacity model optimized through a unified workflow can directly produce high-quality, fully rendered posters without modular design. This enables holistic integration of textual content, visual elements, and layout considerations, while leveraging vision-language feedback during inference to achieve greater coherence and heightened aesthetic appeal.
Unlike prior methods that impose explicit layout embeddings~\cite{zheng2023layoutdiffusion,gao2025postermaker,textdiffuser} or depend on external VLM-based designers~\cite{chen2025posta,yang2024posterllava,seol2024posterllama} (i.e. inherently restrict a model’s expressive freedom), we unlock the potential of a standard diffusion backbone via comprehensive workflow optimization rather than intricate architectural modifications.
Therefore, our paradigm remains fully compatible with existing techniques and offers a flexible foundation for future advances. Fig.\ref{overview} illustrates our unified optimization workflow, and Fig.\ref{dataset fig} details the dataset construction pipeline supporting each stage in our unified framework.

\vspace{-0.1cm}

\subsection{Scalable Text Rendering Optimization}
In the first stage of our workflow, we target the challenge of accurate text rendering, a persistent bottleneck in poster generation. Progress is hindered by two factors: (i) the scarcity of large-scale, high-quality datasets with perfectly rendered text, and (ii) most available text datasets feature plain or low-quality backgrounds, which easily make the model lose the ability to represent common backgrounds.
To overcome these issues, we construct Text-Render-2M via an automated pipeline, producing 2 million samples with diverse text (varying in content, size, count, placement, and rotation) rendered crisply onto high-quality backgrounds. Each text instance is paired with precise captions merged seamlessly with existing image captions.
This dataset ensures both 100\% text rendering accuracy and rich background diversity, enhancing fidelity and robustness in real-world scenarios.
Fig.\ref{dataset fig} illustrates Text-Render-2M, with construction details and examples provided in the supplementary.
We then fine-tune foundation models on paired Text-Render-2M using the flow matching loss~\cite{flux,sd3} to enhance text rendering:

\begin{equation}
\mathcal{L}^{\rm text}_{\rm flow}(\phi)
=
\mathbb{E}_{t\sim\mathcal{U}(0,1),\,x_0,\,\varepsilon}
\bigl\|\,v_\phi(x_t, t)-\dot{x}_t\bigr\|_2^2,
\end{equation}
where $x_t=\alpha_t x_0+\sigma_t\varepsilon$ follows the forward noising schedule, $\dot{x}_t$ is its time derivative, and $v_\phi$ predicts the velocity field. Discretized over timesteps, this loss encourages the model to match the true data flow and yields markedly improved text rendering. As shown in Fig.\ref{text rendering effect}, the model augmented with our text‐rendering optimization achieves significant gains in both rendering accuracy and text alignment across poster typography, plain‐text scenes, and long‐text posters.

\subsection{High-quality Poster Fine-tuning}
\noindent{\textbf{HQ-Poster-100K.}} To develop a high-quality dataset for supervised fine-tuning, we introduce HQ-Poster-100K, a meticulously filtered poster dataset. Our filtering pipeline begins with MD5 hash calculations to eliminate exact duplicates. To avoid including posters with extensive credit/billing blocks, we implement an MLLM-based scoring system. This system presents a binary choice question regarding the presence of credit/billing blocks, with the model outputting logits that are transformed through Softmax to obtain probabilistic scores. We calculate the score based on the following formula, where scores closer to 1 indicate better alignment with our criteria:


\begin{equation}
prob_x = \frac{e^{l_x}}{\sum_{x \in L} e^{l_x}}, 
\qquad
score = \sum_{x \in L} prob_x \cdot w_x 
\end{equation}

Where set $L$ contains all the option letters, $l_x$ represents the logit for option $x$. Here we only consider two-choice questions, that is, the case where there are only $A$ and $B$ in $L$; the weight $w_A$ is set to 0 and $w_B$ is set to 1.
In our filtering pipeline, we utilize InternVL2.5-8B-MPO~\cite{InternVLTeam2024InternVL25} for logits computation with a threshold of 0.98, allowing precise control over filtering stringency. We further refine the dataset using perceptual hashing to remove visually similar posters. The remaining posters are captioned using Gemini2.5-Flash and subjected to HPS scoring, with posters scoring below 0.25 being filtered out to ensure aesthetic quality alignment with human preferences.

To support Region-aware Calibration, HQ-Poster-100K also provides precise text region masks for each poster. Due to the limitations of traditional OCR in recognizing artistic typography, we use Gemini2.5-Flash to extract text region coordinates. Based on their proportional size relative to the poster, masks are classified as Major or Minor, denoting large and small text areas respectively, as shown in Figure\ref{dataset fig}. More details can be found in our supplementary material.

\noindent{\textbf{Region-aware Calibration.}} In poster design, harmony between text and background is crucial. Since our first stage has already improved the text rendering capability of the model, this fine-tuning phase shifts the focus to overall poster style. Therefore, we propose Region-aware Calibration to achieve this purpose. Specifically, essential text carries the core message and is assigned moderate weight to ensure clarity and integration with the background; by contrast, small text—occupying minimal space and prone to rendering errors—is downweighted to prevent distracting collapse. Non-text regions, which define the visual style of the poster, receive full emphasis to guarantee a smooth transition from high-quality imagery to a unified aesthetic layout. This balanced weighting strategy allows the fine-tuned model to preserve text accuracy while strengthening the artistic integrity of the aesthetic poster.

We implement this via a per-pixel weight map $w(p)$:

\begin{equation}
w(p) =
\begin{cases}
0.6 & \text{if } p \in \mathrm{LargeTextMask},\\
0.2 & \text{if } p \in \mathrm{SmallTextMask},\\
1.0 & \text{otherwise},
\end{cases}
\end{equation}
and define our weighted loss as:

\begin{equation}
\mathcal{L}_{\mathrm{flow}}^{\mathrm{poster}}
=
\mathbb{E}_{t\sim\mathcal{U}(0,1),\,x_0,\,\varepsilon}
\Bigl\|\,\bigl(v_{\phi}(x_t,t)-\dot{x}_t\bigr)\,\odot\,w\Bigr\|_2^2,
\end{equation}
where "$\odot$" denotes point-wise multiplication by the weight map $w$. Different from previous scalable text rendering optimization, here we multiply the scaling factor $w$, which encourages the model to learn both crisp text information and a cohesive aesthetic style.
\subsection{Aesthetic-Text Reinforcement Learning}
\noindent{\textbf{Poster-Preference-100K.}}
To further enhance poster generation aesthetics and text rendering capabilities, we develop the Poster-Preference-100K dataset. We use about 20K prompts and generate 5 images for each prompt using the model after Region-aware calibration, for a total of 100K poster images, which is the basic data source for us to construct preference pairs. Using HPSv2~\cite{wu2023human}, we evaluate 5 generated posters per group for human preference scoring, selecting the highest-scoring poster as the preferred sample and the lowest-scoring one as the rejected sample to form preference pairs. Since HPSv2 only evaluates image content and aesthetics, we employ Gemini2.5-Flash to verify text rendering accuracy and style consistency with captions in the preferred posters, filtering out inconsistent pairs. This process yields 6K preference pairs meeting two criteria: 1) an HPSv2 score difference exceeding 0.025, and 2) complete text accuracy in preferred posters.

\noindent{\textbf{Aesthetic–Text Preference Optimization.}}
While the preceding stages guarantee pixel-level text fidelity and precisely calibrated poster styles, they fail to capture the higher-order trade-offs that render a poster genuinely compelling. In particular, (i) detailed preferences, such as subtle layout balance, color harmony, and typographic cohesion, which require global evaluation beyond per-pixel accuracy; (ii) even after achieving crisp text rendering, further corrective tuning is necessary to alleviate residual errors and seamlessly integrate text with the holistic aesthetic. To address these gaps, we frame poster generation as a reinforcement learning problem in this stage: the model must not only denoise accurately but also preferentially generate outputs that satisfy holistic aesthetic criteria.

Concretely, for each prompt, we sample $n$ poster variants $\{x^{(i)}\}_{i=1}^n$ under the current diffusion policy and collapse them into a single “winning” and “losing” pair via best-of-$n$ selection under the combined aesthetic–text reward $R(x)$:

\begin{equation}
x^+ = \arg\max_i R\bigl(x^{(i)}\bigr),
\qquad
x^- = \arg\min_i R\bigl(x^{(i)}\bigr).
\end{equation}

We then optimize the Direct Preference Optimization (DPO)~\cite{dpo} objective:
\begin{equation}
\mathcal{L}_{\rm RL}(\theta)
=
-\,\mathbb{E}_{c}\Bigl[\log\sigma\!\Bigl(\beta\Bigl(\log\frac{p_\theta(x^+\!\mid c)}{p_{\rm ref}(x^+\!\mid c)}
-\log\frac{p_\theta(x^-\!\mid c)}{p_{\rm ref}(x^-\!\mid c)}\Bigr)\Bigr)\Bigr],
\end{equation}
where $p_{\rm ref}$ denotes the fixed reference distribution, and $p_\theta$ is learned diffusion policy parameterized by $\theta$. Because the marginal $p_\theta(x_0\mid c)$ is intractable, we employ the ELBO over the full diffusion chain to evaluate these log-ratio rewards, following prior work~\cite{diffusiondpo,wang2025designdiffusion}. In this way, best-of-$n$ Aesthetic–Text Preference Optimization directly injects a unified preference signal into the diffusion training process.

\vspace{-0.1cm}
\subsection{Vision-language Feedback Refinement.}
\noindent{\textbf{Poster-Reflect-120K.}}
To address potential deficiencies in content and aesthetic quality of initially generated posters, we implement reflection optimization to enhance accuracy and aesthetic value. We introduce the Poster-Reflect-120K dataset, generating six posters for each prompt using our preference-learned model, totaling 120K generated poster images. Subsequently, we employ Gemini2.5-Flash to select the optimal poster from each set of six as the target image for feedback learning. This target image is required to have optimal prompt alignment, superior aesthetic value, and fully correct text rendering. This process yields 5 reflection-pairs from each set of 6 generated images.

During the feedback collection phase, we gather suggestions in two key areas: \textbf{\textit{Poster Content Suggestions}} and \textbf{\textit{Aesthetic Style Optimization Suggestions}}, with Gemini2.5-Flash analyzing both the target poster and the poster requiring optimization to provide comprehensive feedback. To optimize our prompting strategy for both feedback and VLM fine-tuning, we implement specific guidelines: the model is instructed to perform internal comparisons without explicitly referencing the second reference poster, and feedback is structured as concrete editing instructions.

\noindent{\textbf{Reflect VLM fine-tuning.}}
To obtain optimization feedback during inference, we construct VQA samples by embedding the original caption in the prompt alongside the generated poster requiring optimization, and using Gemini2.5-Flash-generated feedback as supervision. Specifically, this input configuration maintains consistency between training and inference phases, excluding reference target posters in both cases and using the original prompt as the baseline for suggestions. Additionally, when generating feedback with Gemini2.5-Flash, we deliberately utilize only target posters as references, omitting original captions to preserve model creativity.

\noindent{\textbf{Joint Vision–Language Conditioning.}} 
For the poster construction, iterative critique—combining visual inspection with targeted verbal feedback—is essential for refining both aesthetic content and background harmony. Inspired by this, we introduce a joint vision–language feedback loop for multimodal corrections in a unified workflow. For each generated–ground‐truth pair, Gemini produces two textual reflections, $f_{c}$ (Poster Content Suggestions) and $f_{s}$ (Aesthetic Style Suggestions). Rather than appending these strings to the original prompt—which would exceed the encoder’s length and degrade performance—we jointly encode them via a text encoder $E_{t}$, yielding $e_{c,s}=E_{t}(f_{c},f_{s})$, and then concatenate this with the original prompt embedding $e_{p}$ (with positional encodings to preserve order and semantics).
Additionally, drawing on OmniControl \cite{tan2024ominicontrol}, we inject the image‐level feedback signal $v_{\rm img}$ (encoded by VAE) directly into the conditioning branch. The resulting multimodal context is:
\begin{equation}
c = [\,e_{p};\,e_{c,s};\,v_{\rm img}\,]
\end{equation}
where $c$ serves as the conditioning input. Finally, we fine-tune the model under the conditional flow matching objective:

\begin{equation}
\mathcal{L}^{VL}_{\rm flow}(\theta)=\mathbb{E}_{t,x_0,\varepsilon}\|v_{\theta}(x_t,t\mid c)-\dot x_t\|_2^2,
\end{equation}
which enables the model to iteratively refine its outputs in response to structured textual reflections and semantically enriched visual feedback.
\vspace{-0.1cm}
\subsection{Inference}
During inference, PosterCraft generates a complete poster from a single user prompt. The prompt is first processed by a MLLM (e.g., Qwen3-based~\cite{yang2025qwen3} Magic Prompter), which enriches the input with detailed aesthetic cues to guide layout and content generation. PosterCraft then produces the poster in an end-to-end manner, without relying on additional inputs or modular designs.
Additionally, an VLM-based critique loop can further enhance the result by providing structured multi-modal feedback. This enables iterative refinement to improve aesthetic quality and semantic alignment.

\begin{figure}[!t]
    \centering
    \setlength{\abovecaptionskip}{0cm} 
    \setlength{\belowcaptionskip}{-0cm}
    \includegraphics[width=8.5cm]{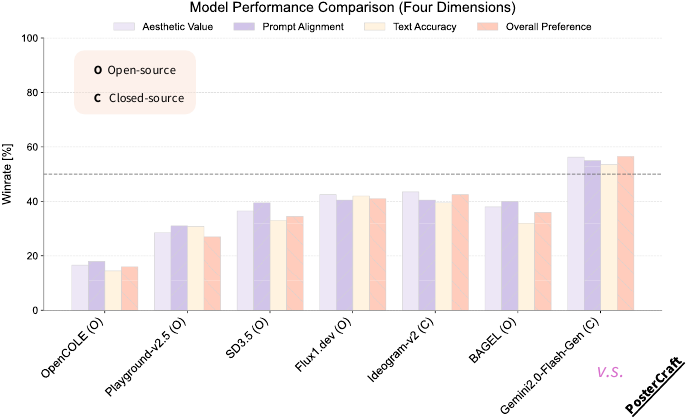}
    \caption{\textbf{User study comparisons} between PosterCraft and both SOTA open-source and closed-source models. PosterCraft consistently outperforms all open-source baselines and several proprietary systems cross multiple dimensions, with performance marginally below that of the leading closed-source model, Gemini2.0-Flash-Gen.}
    \label{fig:user study}
    \vspace{-0.3cm}
\end{figure}

\begin{figure*}[!t]
    \centering
    \setlength{\abovecaptionskip}{0cm} 
    \setlength{\belowcaptionskip}{-0cm}
    \includegraphics[width=17.5cm]{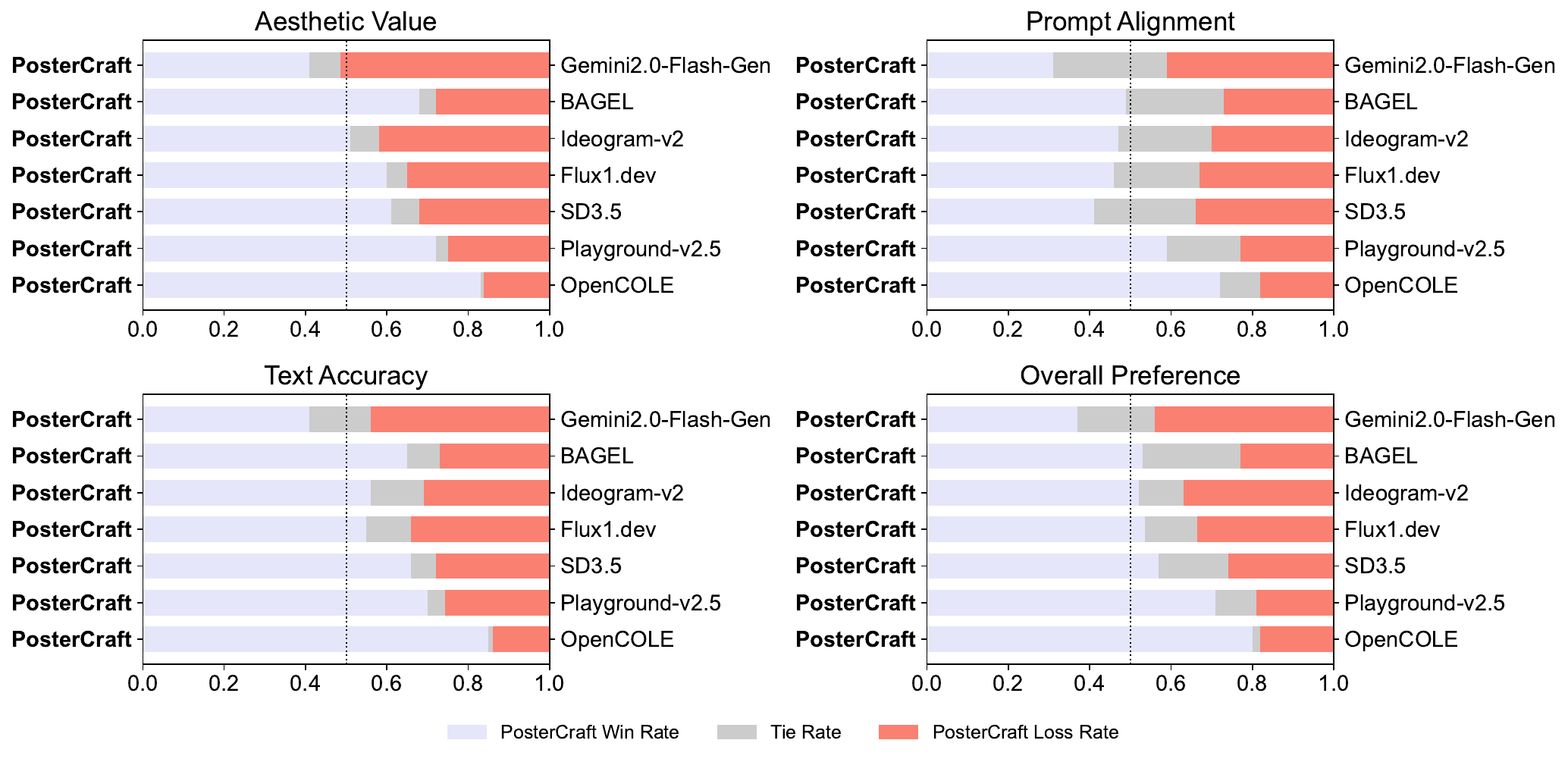}
    \caption{\textbf{Gemini serves as an authoritative evaluator for human preference comparisons} across our method and other baselines. PosterCraft outperforms most state-of-the-art generative models in aesthetic coherence, prompt alignment, text rendering, and overall preference. It achieves performance nearly on par with the leading commercial model Gemini2.0-Flash-Gen in text rendering, while showing only a slight gap in other aspects.}
    \label{fig:gemini}
    \vspace{-0.2cm}
\end{figure*}

\begin{figure*}[!t]
    \centering
    \setlength{\abovecaptionskip}{0.0cm} 
    \setlength{\belowcaptionskip}{-0cm}
    \includegraphics[width=17.5cm]{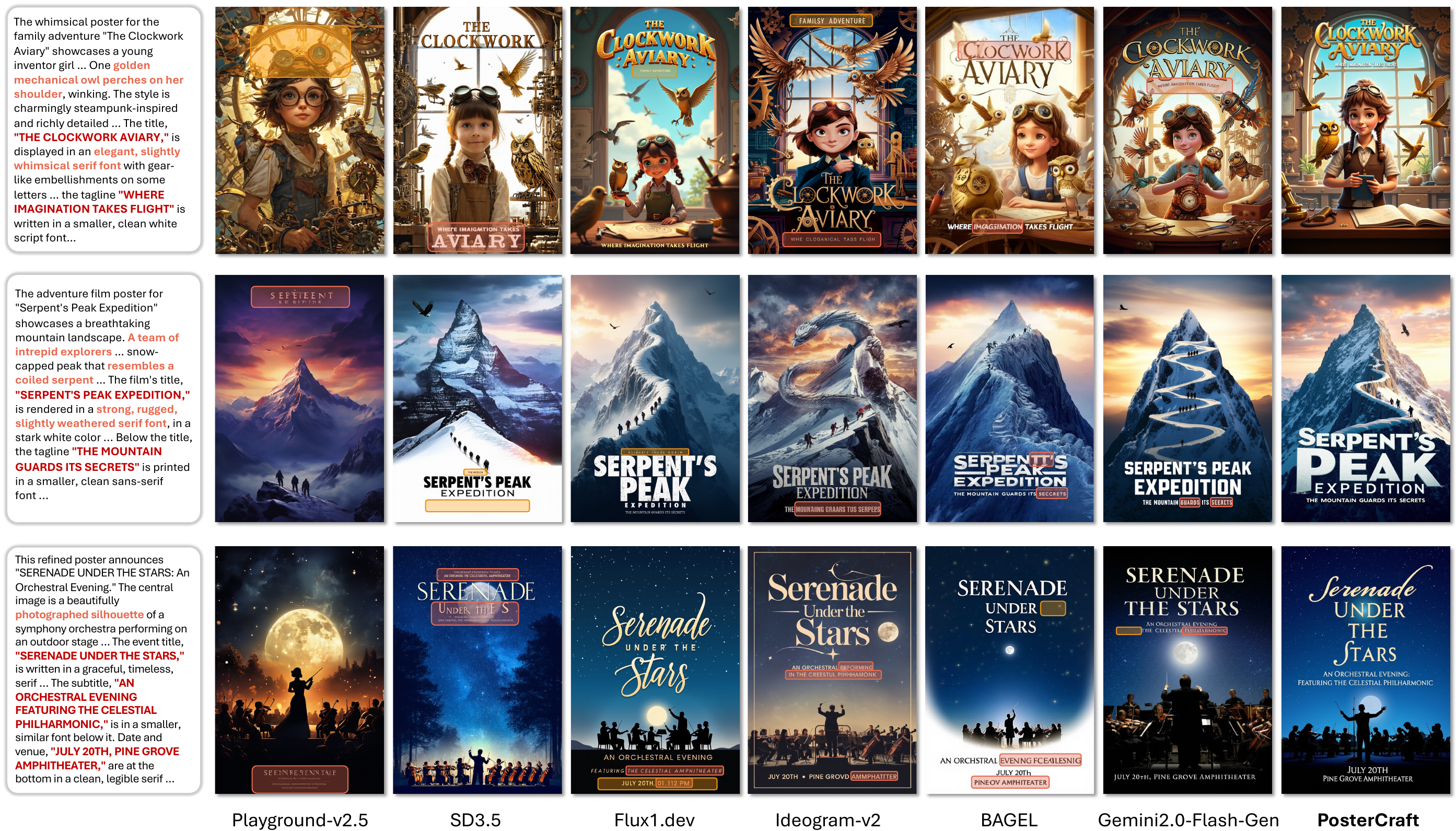}
    \caption{\textbf{Visual comparison} of different model outputs. \colorbox{lred}{Red boxes} highlight misspelled or distorted text, while \colorbox{lyellow}{yellow boxes} indicate redundant or missing text elements. Within the prompts, \textcolor{orange}{orange text} denotes content and style requirements, and \textcolor{red}{red text} indicates textual elements. From the visual comparison, it is evident that our method achieves superior aesthetic appeal compared to existing state-of-the-art approaches. In terms of text presentation, our rendered fonts blend more naturally with the scene content, and text rendering errors are nearly eliminated.}
    \label{fig:visual}
    \vspace{-0.1cm}
\end{figure*}

\begin{figure*}[!t]
    \centering
    \setlength{\abovecaptionskip}{0.0cm} 

    \setlength{\belowcaptionskip}{0.cm}
    \includegraphics[width=17.5cm]{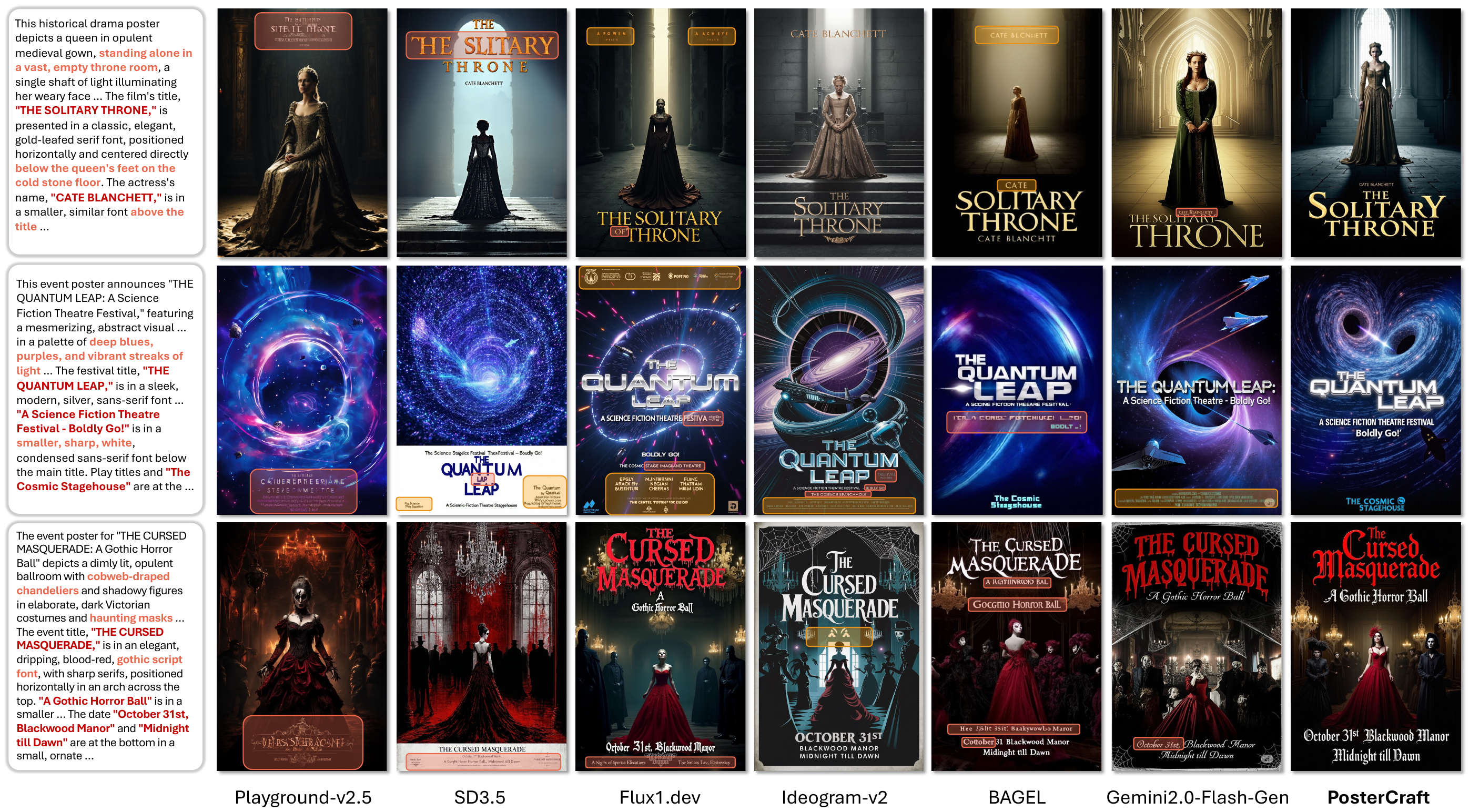}
    \caption{\textbf{Visual comparison} of different model outputs. \colorbox{lred}{Red boxes} highlight misspelled or distorted text, while \colorbox{lyellow}{yellow boxes} indicate redundant or missing text elements. Within the prompts, \textcolor{orange}{orange text} denotes content and style requirements, and \textcolor{red}{red text} indicates textual elements. It indicates that our method significantly outperforms existing SOTA approaches in generating high-quality posters under long-prompt conditions, with notably improved prompt alignment. In terms of text rendering, our model produces fonts that align closely with the visual context of the scene, with minimal rendering errors.}

    \label{fig:visual_2}
    \vspace{-0.3cm}
\end{figure*}

\vspace{-0.2cm}
\section{Experiments}
\subsection{Implementation}
For PosterCraft, we initialize from the Flux-dev~\cite{flux} backbone and train exclusively in mixed precision. In text rendering optimization, we perform 300 K full-parameter iterations on Text-Render-2M using Adafactor~\cite{shazeer2018adafactor} with a small learning rate of 2e-6. Stage 2 consists of 6,000 full-parameter fine-tuning steps on HQ-Poster-100K—again with Adafactor at lr = 1e-5 —during which we apply per-pixel flow-matching weights of {0.6, 0.2, 1.0} in different regions. In reinforcement learning, for each prompt we sample n=5 candidates and optimize our reinforcement learning objective through AdamW~\cite{adamw} (lr = 1e-4) in 1500 steps, fine-tuning only LoRA adapters with rank 64. For vision-language feedback refinement, we encode the dual-language reflections $f_c$ and $f_s$ via T5~\cite{t5}. Then we fine-tune the LoRA adapters (rank 128) under the conditional flow-matching loss for 6000 steps using AdamW (lr = 1e-4). We employ Internvl3-8B~\cite{InternVLTeam2025InternVL3} as our feedback generation model, conducting fine-tuning for 2 epochs and setting the temperature to 0 during inference.
\vspace{-0.1cm}

\begin{figure*}[!t]
    \centering
    \setlength{\abovecaptionskip}{0.0cm} 
    \setlength{\belowcaptionskip}{-0cm}
    \includegraphics[width=17.5cm]{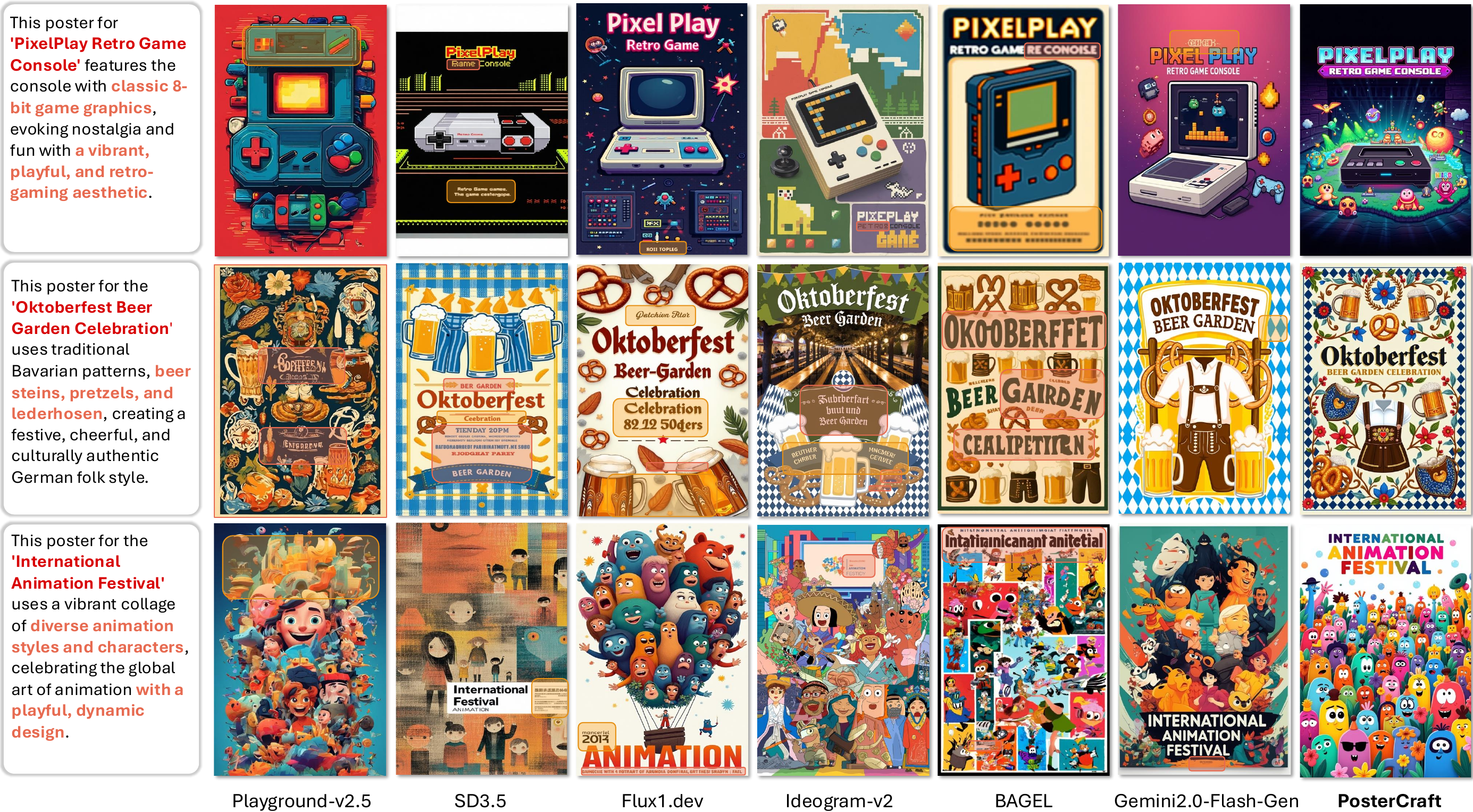}
    \caption{\textbf{Visual comparison} of different model outputs. \colorbox{lred}{Red boxes} highlight misspelled or distorted text, while \colorbox{lyellow}{yellow boxes} indicate redundant or missing text elements. Within the prompts, \textcolor{orange}{orange text} denotes content and style requirements, and \textcolor{red}{red text} indicates textual elements. Compared to other methods, our approach produces cleaner layouts, better theme alignment, and more accurate text rendering under short prompts.}
    \label{fig:visual_3}
    \vspace{-0.2cm}
\end{figure*}

\begin{figure*}[!t]
    \centering
    \setlength{\abovecaptionskip}{0.1cm} 
    \setlength{\belowcaptionskip}{-0cm}
    \includegraphics[width=17.5cm]{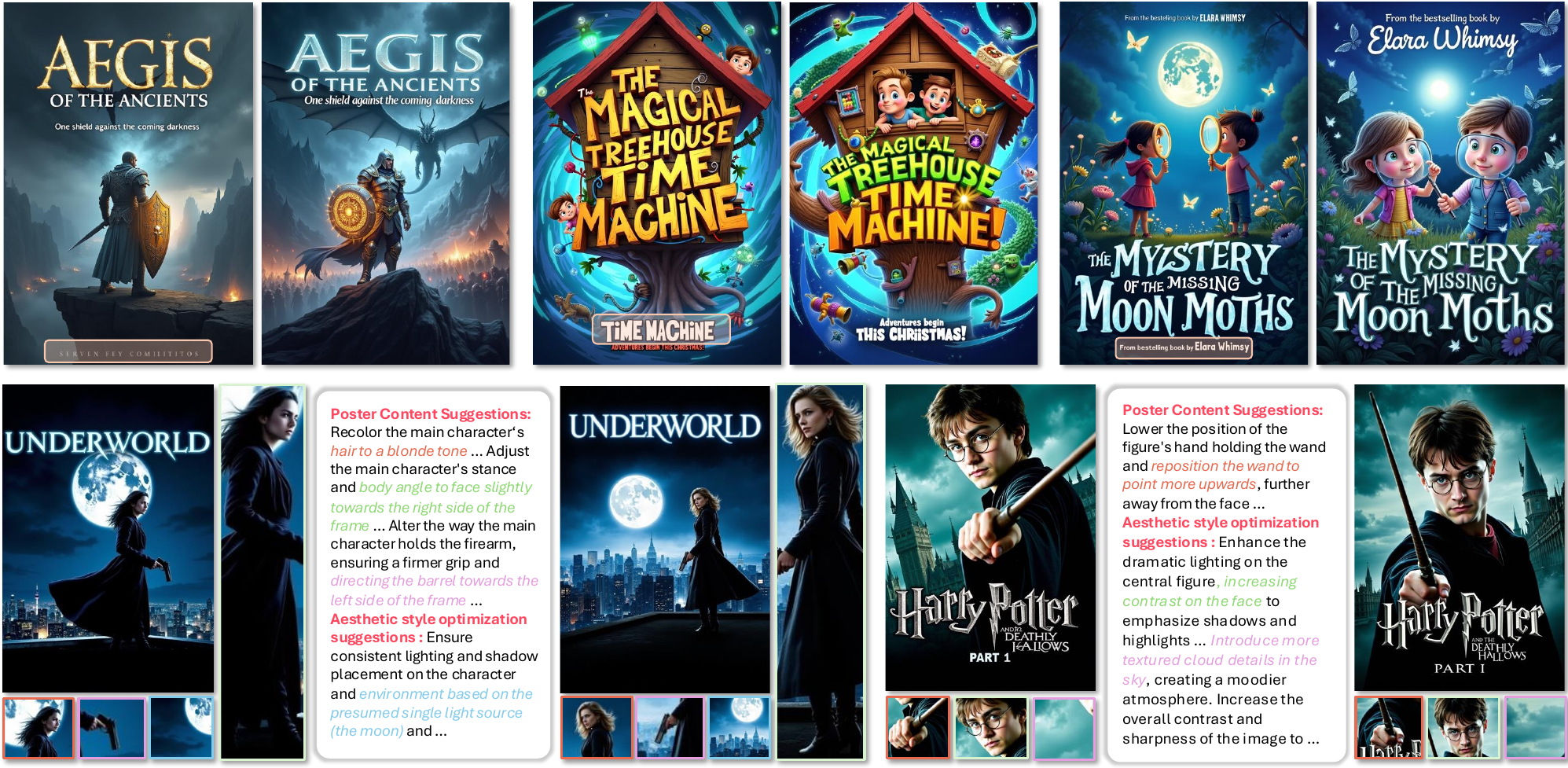}
    \caption{\textbf{Aesthetic-text reinforcement learning (top row) and vision-language feedback (bottom row) qualitative comparisons.} \colorbox{orange}{orange boxes} denote the text biases. Different color fonts represent key feedback information from VLM. The top examples demonstrate that our reinforcement learning stage effectively improves overall aesthetic quality and partially corrects text rendering errors after supervised fine-tuning. The bottom examples illustrate the impact of vision-language reflection, where feedback from a VLM is integrated into the generation loop. This results in noticeable enhancements to both visual aesthetics and semantic coherence in the final poster outputs.}
    \label{dpo and reflection}
    \vspace{-0.1cm}
\end{figure*}

\vspace{-0.1cm}
\subsection{Quantitative Results and Comparisons}
We conduct a quantitative comparison of our PosterCraft against seven leading open-source advanced closed-source commercial models. To assemble our test prompts, we use Gemini2.0-Flash-Gen~\cite{gemini} to randomly generate 100 aesthetic poster prompts—balanced across short, medium, and long lengths—and for each prompt we sample three output per model, yielding 300 test images. We generate posters with OpenCOLE~\cite{inoue2024opencole}, Playground-v2.5~\cite{liu2024playground}, SD3.5~\cite{stabilityai2024sd35large}, Flux1.dev~\cite{flux}, Ideogram-v2~\cite{ideo}, BAGEL~\cite{bagel}, and Gemini2.0-Flash-Gen~\cite{gemini}, apply the OCR engine of the SOTA VLM~\cite{gemini} to each image, and report three precision-oriented metrics—text recall, text F-score, and text accuracy—averaged across all 300 samples.

As shown in Tab.\ref{tab:text_metrics}, PosterCraft achieves substantially higher recall and F1 scores, capturing a more complete character set, while its accuracy surpasses leading open-source methods (Flux1.dev, SD3.5) and competitive commercial solutions (Ideogram-v2). PosterCraft's performance closely approaches that of mature SOTA commercial systems such as Gemini2.0-Flash-Gen, demonstrating that its text-rendering not only advances academic benchmarks but also offers near production-level competitiveness.

Besides, we conduct the user study and a Gemini-based evaluation to assess model quality.
Twenty experienced poster designers evaluate outputs across multiple dimensions, while Gemini2.0-Flash serves as a strict evaluator via carefully designed prompts, as shown in Fig.\ref{fig:user study} and Fig.\ref{fig:gemini}.
Both evaluations show that our method significantly enhances aesthetics and text rendering over the base model Flux1.dev, and outperforms all state-of-the-art open-source and several closed-source models across all evaluation metrics. Additionally, PosterCraft performs only slightly below Gemini2.0-Flash-Gen, validating the effectiveness of our unified workflow in fully unlocking the generative potential of powerful baseline model.

\vspace{-0.1cm}
\subsection{Qualitative Results and Comparisons}
To further evaluate the superiority of vision, we conduct a comprehensive visual comparison between PosterCraft and six other models—four open-source and two proprietary commercial systems, as shown in Fig.\ref{fig:visual}, Fig.\ref{fig:visual_2} and Fig.\ref{fig:visual_3}. Playground-v2.5 exhibits complete failure in text rendering, while SD3.5~\cite{stabilityai2024sd35large} achieves only partial title rendering, with both models showing deficiencies in aesthetic quality and prompt adherence, particularly in representing animated styles and semantics. Although Flux1.dev\cite{flux} and BAGEL~\cite{bagel} demonstrate improved title rendering capabilities, they still contain textual errors and fall short in aesthetic quality and prompt alignment, failing to incorporate specific details such as "One golden mechanical owl perches on her shoulder." Among proprietary models, Ideogram-v2~\cite{ideo} and Gemini2.0-Flash-Gen~\cite{gemini} show minor text errors only in smaller text elements and achieve superior aesthetic results. However, both demonstrate prompt adherence issues: Ideogram-v2 fails to accurately depict mountain formations, Gemini lacks photorealism, and neither successfully generates silhouette effects in the third image.

In the short-prompt setting of Fig.\ref{fig:visual_3}, PosterCraft maintains a strong balance between visual appeal and accurate text rendering. It consistently integrates title and scene elements, for example, seamlessly embedding 'PixelPlay Retro Game Console' or  'International Animation Festival' in stylized compositions. Competing models often display legibility issues (e.g., SD3.5, Flux1.dev), omit key poster elements (e.g., BAGEL, Ideogram-v2), or suffer from text-scene disconnection (e.g., Playground-v2.5). While Gemini2.0-Flash-Gen performs well in text rendering, it still suffers from aesthetic limitations, often producing visually monotonous outputs with missing or underdeveloped design elements. Our results stand out with their vibrant layout, theme adherence, and natural visual-text coherence, even under minimal input conditions.

\begin{table}[t]
  \centering
  \setlength{\abovecaptionskip}{0.0cm} 
\setlength{\belowcaptionskip}{-0.0cm}
  \caption{\textbf{A text quality comparison} with SOTA poster generation models, demonstrates that PosterCraft achieves superior performance across recall, F-score, and accuracy, while only slightly below the recent closed-source Gemini2.0-Flash-Gen. We highlight the \colorbox{best}{best} and \colorbox{second}{second} metrics.
\textcolor{gray_venue}{Open} and \textcolor{gray_venue}{Close} denote open-source and closed-source.}
  \resizebox{8.5cm}{!}{
  \begin{tabular}{ l c c c }
    \toprule
    \textbf{Method} & \textbf{Text Recall ↑} & \textbf{Text F-score ↑} & \textbf{Text Accuracy ↑} \\
    \midrule
    OpenCOLE~\cite{inoue2024opencole} \textcolor{gray_venue}{\small{(Open)}}     & 0.082 & 0.076  & 0.061  \\
    Playground-v2.5~\cite{yang2024posterllava}  \textcolor{gray_venue}{\small{(Open)}}  & 0.157 & 0.146  & 0.132  \\
    SD3.5~\cite{stabilityai2024sd35large}  \textcolor{gray_venue}{\small{(Open)}}        & 0.565 & 0.542  & 0.497  \\
    Flux1.dev~\cite{flux}  \textcolor{gray_venue}{\small{(Open)}}    & 0.723 & 0.707  & 0.667  \\
    Ideogram-v2~\cite{ideo} \textcolor{gray_venue}{\small{(Close)}} & 0.711 & 0.685  & 0.680  \\
    BAGEL~\cite{bagel} \textcolor{gray_venue}{\small{(Open)}} & 0.543 & 0.536  & 0.463  \\
    Gemini2.0-Flash-Gen~\cite{gemini} \textcolor{gray_venue}{\small{(Close)}}   & \colorbox{best}{0.798} & \colorbox{best}{0.786} & \colorbox{best}{0.746} \\
    \midrule
    \textbf{PosterCraft (ours)} 
                   & \colorbox{second}{0.787}
                   & \colorbox{second}{{0.774}}
                   & \colorbox{second}{{0.735}} \\
    \bottomrule
  \end{tabular}}  \label{tab:text_metrics}
     \vspace{-0.5cm}
\end{table}

\vspace{-0.1cm}
\section{Ablation Study}
In this section, we validate the efficacy of our proposed workflow by independently assessing the contribution of each stage. To ensure a fair comparison, all parameter settings and experimental conditions are held identical to those used in the preceding experiments.

\noindent{\textbf{Text Rendering Optimization is critical for both accuracy and perception.}}
Text Rendering Optimization ensures clarity and fidelity.
Its removal leads to drops in both OCR accuracy and human preference, confirming that large-scale, high-quality text data significantly improves text rendering. The diverse and realistic backgrounds also preserve visual quality, while models without this optimization often fail to maintain legibility and accuracy, as shown in Fig.\ref{text rendering effect}.
\noindent{\textbf{Region-aware Calibration improves poster consistency.}}
It helps the model adapt to spatial context and balance text-background. Without it, all regions are treated equally, weakening stylistic coherence in visually complex posters while suffering text bias.
\noindent{\textbf{Aesthetic-Text Reinforcement Learning boosts visual appeal.}}
RL strongly impacts human preference, showing its importance in learning higher-order aesthetic cues like layout balance, color harmony, and typographic cohesion. It makes outputs more visually compelling.
The top-row results in Fig.\ref{dpo and reflection} indicate that reinforcement learning strongly impacts holistic preference, highlighting its effectiveness in capturing higher-order aesthetic cues and text rendering. This leads to more visually compelling outputs, with clear improvements evident in the poster compositions across the top row.
\noindent{\textbf{Reflection refines outputs with vision-language feedback.}}
Multi-modal reflection is beneficial for high-level semantic correction. Reflection enables the model to assess and iteratively improve generation quality, particularly in terms of stylistic integration. As shown in the bottom row of Fig.\ref{dpo and reflection}, vision-language reflection can effectively refine the overall aesthetic appeal and layout consistency.
These results validate the necessity of our workflow components and show that unified, targeted optimization can unlock the full potential of powerful foundation models for end-to-end aesthetic poster generation.usion models for end-to-end aesthetic poster generation.

\begin{figure}[!t]
    \centering
    \setlength{\abovecaptionskip}{0cm} 
    \setlength{\belowcaptionskip}{-0.2cm}
    \includegraphics[width=8.3cm]{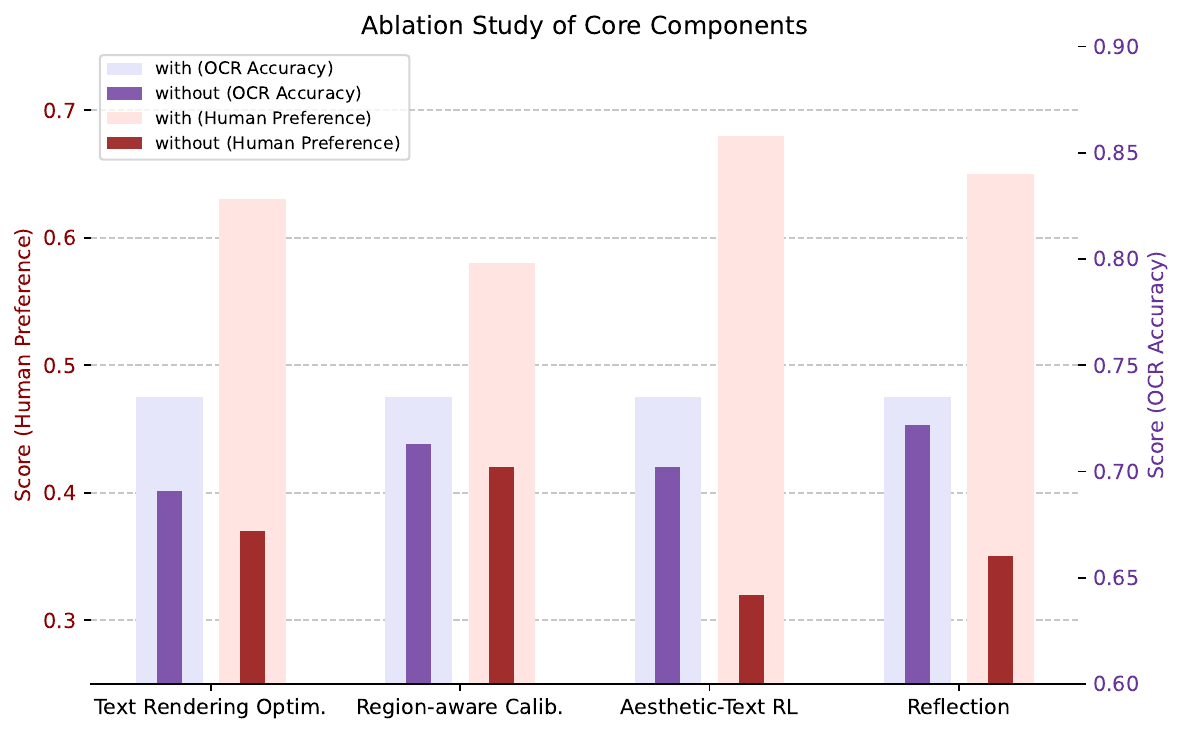}
    \caption{\textbf{Ablation experiments on the core components of our workflow}: removing Text Rendering Optimization, Region-aware Calibration, Aesthetic-Text RL, or Reflection leads to sustained declines in OCR accuracy (purple) and human preference (peach), demonstrating the effectiveness of our overall design, targeted optimizations and validating our motivation.}
    \label{fig:ab}
    \vspace{-0.2cm}
\end{figure}
\section{Conclusion}
PosterCraft presents a unified, cascaded workflow that integrates scalable text-rendering optimization, poster fine-tuning, RL-driven aesthetic enhancement, and joint vision–language feedback, demonstrating powerful foundation model can directly produce posters with both striking visuals and precise text. Our fully automated dataset pipelines support scalable, task-specific training, and we achieve substantial gains over open-source baselines, approaching leading commercial quality.

\bibliographystyle{ACM-Reference-Format}
\bibliography{sample-base}

\clearpage

This is supplementary material for \textbf{\textit{PosterCraft: Rethinking High-Quality Aesthetic Poster Generation in a
Unified Framework.}} 

We present the following materials in this supplementary material:

\begin{itemize}
    \item Sec.\ref{text data} Detailed dataset construction and comparison on Text-Render-2M.\\
    \item Sec.\ref{poster100k data} More information about the automatic processing pipeline for HQ-Poster-100K.\\
    \item Sec.\ref{dpo data} More examples and explanations of the Poster-Preference-100K.\\
    \item Sec.\ref{reflect data} Detailed illustration of the Poster-Reflect-120K.\\
    \item Sec.\ref{ocr and evaluator} Gemini for OCR calculation and preference evaluator.\\
    \item Sec.\ref{supp visual} Additional visual results generated by our PosterCraft. \\
    \item Sec.\ref{limitations} Limitations.\\
    \item Sec.\ref{future work} Future work.\\
\end{itemize}

\section{Text-Render-2M Construction Pipeline}\label{text data}
In this section, we provide a detailed overview of the automated construction process behind Text-Render-2M, a large-scale synthetic dataset designed to improve text rendering quality in the baseline model. This dataset plays a critical role in the text rendering optimization stage of our workflow. By overlaying diverse textual elements onto high-resolution background images, it allows the model to learn accurate text generation while preserving its ability to represent rich visual content.

\noindent{Multi-Instance Text Rendering}
Each image contains a variable number of independently placed text instances, typically ranging from one to three. This multi-instance setup introduces compositional complexity and better reflects natural layouts found in aesthetic posters.

\noindent{Text Content Generation}
Text content is synthesized using a hybrid strategy:
\begin{itemize}
    \item A majority of the samples are generated using template-based grammars, with phrases constructed from predefined structures filled with curated vocabulary lists.
    \item A smaller portion uses random alphanumeric strings to simulate noisy or unstructured textual inputs.
\end{itemize}
The generator supports rich variations in punctuation, casing , and structure (e.g., both single-word and short-phrase constructions), ensuring linguistic diversity.

\noindent{Font Selection and Style Variability}
Fonts are randomly drawn from a categorized library containing both standard and artistic typefaces. When multiple styles are available, a roughly even split is enforced between classic and stylized fonts. The system filters fonts that do not support lowercase letters to avoid invalid renderings. This selection mechanism ensures visual variability while maintaining text legibility.

\noindent{Layout and Placement Strategy}
Text is positioned using a 3×3 grid-based partitioning scheme (e.g., top-left, center, bottom-right). Before final placement, bounding box collision checks are performed. If an overlap is detected, the system will retry placement with different positions or reduced font sizes, often within 3–5 attempts per instance. This strategy enables dense yet legible arrangements while minimizing visual clutter.

\noindent{Orientation and Alignment}
Orientation options include:
\begin{itemize}
    \item Horizontal,
    \item Vertically rotated (rotated 90 degrees),
    \item Vertically stacked (one character per line).
\end{itemize}
In the horizontal mode, a small proportion of texts receive a random rotation, and a subset of longer text fragments are automatically wrapped across multiple lines. Alignment is randomly selected among left, center, or right justification.

\begin{figure*}[!t]
    \centering
    \setlength{\abovecaptionskip}{0.2cm} 
    \setlength{\belowcaptionskip}{-0cm}
    \includegraphics[width=17.5cm]{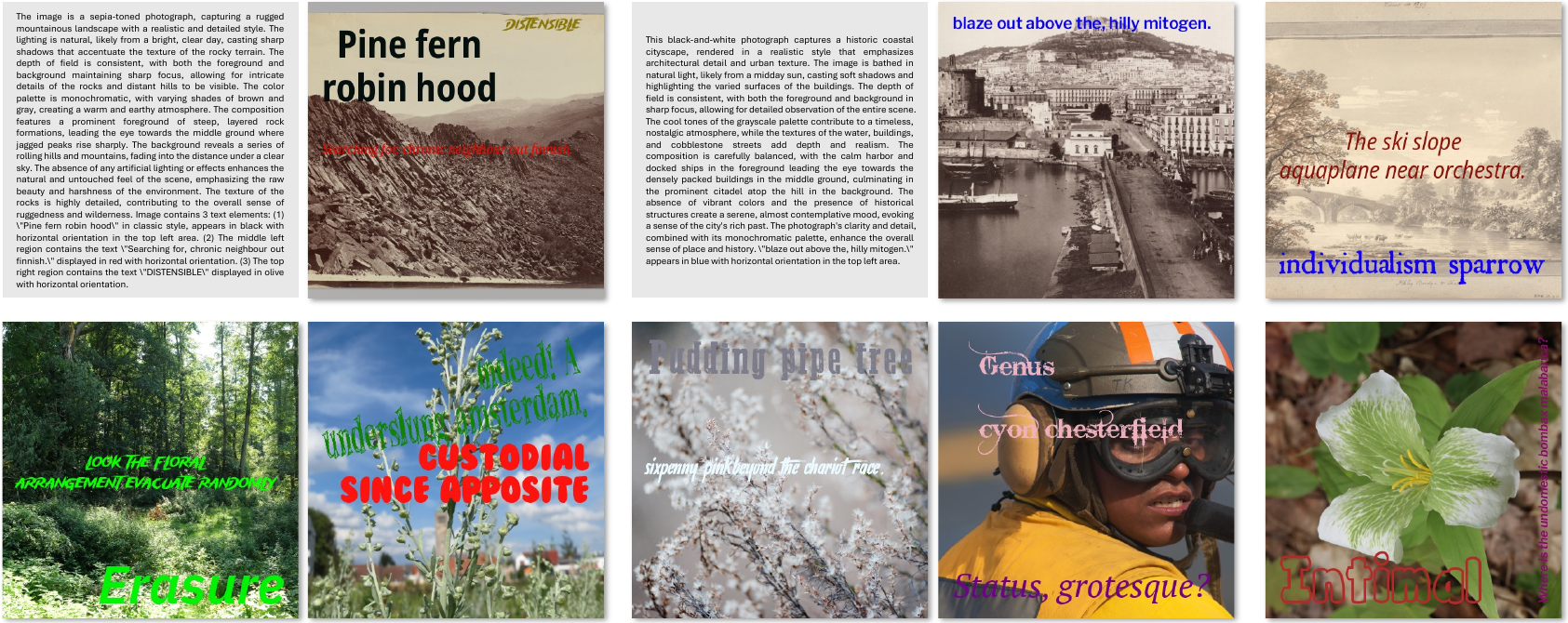}
    \caption{More samples are shown, which are high-quality paired samples in Text-Render-2M.}
    \label{fig:text_render_data}
    \vspace{-0.1cm}
\end{figure*}

\noindent{Prompt Generation}
Each image is paired with a structured natural language prompt. Prompts include: 
\begin{itemize}
    \item The text content,
    \item Position (e.g., “bottom right”),
    \item Orientation (e.g., “vertically stacked”),
    \item Color category, and
    \item Optionally (included in ~50\% of cases), the font style.
\end{itemize}
When multiple texts are rendered, their prompts are numbered and concatenated. If no text is successfully rendered, a fallback prompt indicating the absence of text is generated. We provide a number of samples to view in Fig.\ref{fig:text_render_data}.

\section{Automatic Processing Pipeline for HQ-Poster-100K}\label{poster100k data}
For collecting high-quality poster images, we design an automated image filtering and annotation pipeline for HQ-Poster-100K utilizing three MLLMs:

(1) MLLM Scorer: We employ Internvl2.5-8B-MPO with a binary classification task to filter out posters containing extensive Credit Blocks, Billing Blocks, or "4K ultrahd" cover texts. The complete prompt is shown in Fig.\ref{prompt:2-1-scorer}.

(2) Gemini Caption Generation: We utilize Gemini2.5-Flash-Preview-04-17 for automated poster caption annotation, systematically describing the poster content, visual style, and textual elements. The complete prompt structure is illustrated in Fig.\ref{prompt:2-2-caption}

(3) Gemini Mask Generation: The final step in poster data collection involves generating text region masks. We employ Gemini2.5-Flash-Preview-04-17's robust OCR capabilities to extract text masks from posters, which are subsequently used for Region-aware Calibration. The prompt is shown in Fig.\ref{prompt:2-3-mask}

\begin{figure*}[!t]
    \centering
    \setlength{\abovecaptionskip}{0.2cm} 
    \setlength{\belowcaptionskip}{0cm}
    \includegraphics[width=17.5cm]{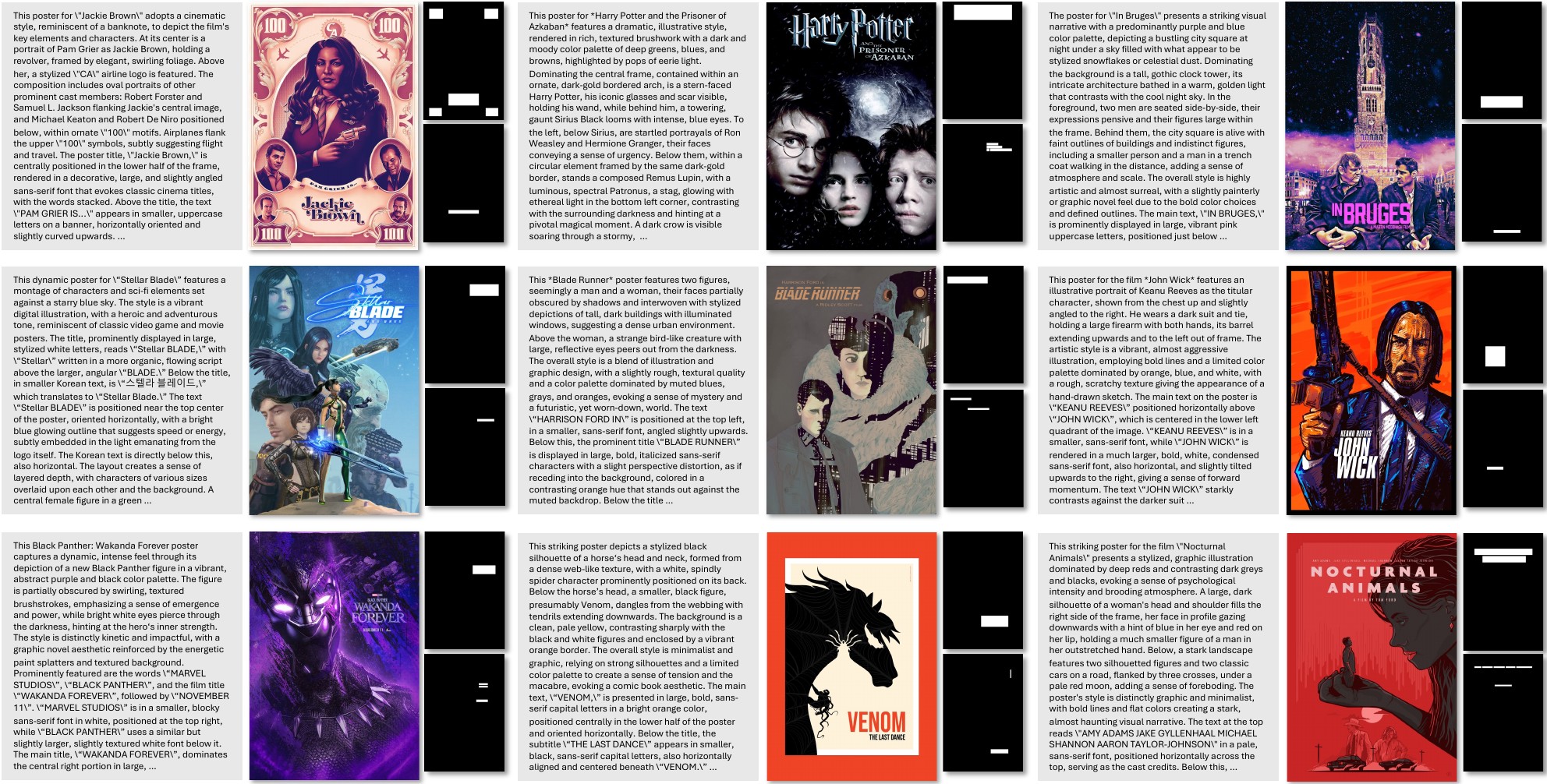}
    \caption{More pictures are shown, which are high-quality paired samples with masks in HQ-Poster-100K.}
    \label{fig:visual_supp_poster100k}
    \vspace{-0.1cm}
\end{figure*}

In addition, we provide more masks, images, prompts triplets in Fig.\ref{fig:visual_supp_poster100k} to demonstrate the advantages of our dataset.

\begin{figure*}[ht]
    \setlength{\abovecaptionskip}{0.1cm} 
    \setlength{\belowcaptionskip}{0cm}
\noindent\begin{example}{MLLM Scorer Prompt}
Does this poster contain a large Billing Block or Credit Block at the bottom or \textit{"4K ultrahd"} text at the top?

Based on your judgment, use the closest option to answer, and only return the label:

A. Yes. There is a large Billing Block or Credit Block at the bottom or \textit{"4K ultrahd"} text at the top.

B. No. There is no Billing Block or Credit Block at the bottom and no \textit{"4K ultrahd"} text at the top.
\end{example}
\caption{\textbf{Prompt} for MLLM Scorer in HQ-Poster-100K.}
\label{prompt:2-1-scorer}
\end{figure*}

\begin{figure*}[ht]
    \setlength{\abovecaptionskip}{0.1cm} 
    \setlength{\belowcaptionskip}{0cm}
\noindent\begin{example}{Gemini Caption Generation}
Please write a structured and detailed caption in a single paragraph for this poster, covering the following five aspects in order:

\textbf{Poster Content}---Describe what is visually depicted.

\textbf{Poster Style}---Describe the visual or artistic tone, such as cinematic, surreal, minimalist, or other distinct aesthetics.

\textbf{Poster Text}---Provide the exact words shown in the image (title, subtitle, slogan, etc.) and their overall communicative intent.

\textbf{Text Style and Position}---Describe the typography in detail, including font style, size, texture, and how it visually blends or contrasts with the background (e.g., carved into a surface, embedded in light, wrapped by natural objects, etc.); also specify where each piece of text is positioned and its orientation angle in the frame.

\textbf{Layout}---Describe how the all elements are arranged to guide the viewer's focus.

Be specific, descriptive, and cohesive. Keep the response between 200 and 300 words, written as a single paragraph. Avoid listing or enumeration. Do not mention any design tools or generation methods. Write as if for a professional design catalog, highlighting how visual and typographic design choices form a unified and compelling narrative.
\end{example}
\caption{\textbf{Prompt} for Gemini Caption Generation in HQ-Poster-100K.}
\label{prompt:2-2-caption}
\end{figure*}

\lstset{
  basicstyle=\ttfamily\small,
  breaklines=true,
  showstringspaces=false,
  commentstyle=\color{gray!70!white},
  keywordstyle=\color{blue!70!black},
  stringstyle=\color{red!70!black},
  numberstyle=\tiny\color{gray},
  numbers=none,
  frame=none,
  rulesepcolor=\color{gray},
  aboveskip=1em,
  belowskip=1em,
  language=json,
}

\begin{figure*}[ht]
    \setlength{\abovecaptionskip}{0.1cm} 
    \setlength{\belowcaptionskip}{0cm}
\begin{example}{Gemini Mask Generation}
Detect all text regions in the image. For each text region, provide its bounding box in \texttt{box\_2d} format \texttt{[ymin, xmin, ymax, xmax]}.
The coordinates for each bounding box must be a list of four integers \texttt{[ymin, xmin, ymax, xmax]}, normalized to the range \texttt{[0, 1000]}. Ensure the box completely covers the text area.

\textbf{MANDATORY GUIDELINES:}
\begin{itemize}
    \item The \texttt{box\_2d} coordinates \texttt{[ymin, xmin, ymax, xmax]} should be integers normalized to 0-1000.
    \item If no text is found in the image, the \texttt{"text\_regions"} list in the JSON output should be empty.
\end{itemize}

\textbf{STRICT CONSTRAINTS:}
\begin{itemize}
    \item Adhere strictly to the JSON output format specified below.
    \item Do not include any explanations, apologies, or conversational text outside of the JSON structure.
    \item Ensure the provided normalized coordinates are accurate.
\end{itemize}

\textbf{RESPONSE FORMAT:}
\begin{itemize}
    \item Respond with a single JSON object. Do NOT use markdown (e.g., \texttt{\textasciigrave\textasciigrave\textasciigrave json ... \textasciigrave\textasciigrave\textasciigrave}).
    \item The JSON object must have a single key \texttt{"text\_regions"}.
    \item The value of \texttt{"text\_regions"} must be a list of \texttt{bounding\_boxes}.
    \item Each \texttt{bounding\_box} must be a list of four integer coordinates \texttt{[ymin, xmin, ymax, xmax]}, normalized to \texttt{[0, 1000]}.
    \item Example of the required JSON structure for \texttt{"text\_regions"} containing two bounding boxes:
        \begin{lstlisting}[language=json]
[
  [ymin1, xmin1, ymax1, xmax1],
  [ymin2, xmin2, ymax2, xmax2]
]
        \end{lstlisting}
    \item The complete JSON object should look like this:
        \begin{lstlisting}[language=json]
{
  "text_regions": [
    // List of bounding_boxes as shown above
    // e.g., [[20, 10, 50, 100], [70, 150, 100, 250]]
  ]
}
        \end{lstlisting}
    \item If no text is found, the output should be: \texttt{\{"text\_regions": []\}}.
    \item Provide ONLY this JSON object.
\end{itemize}

Now, based SOLELY on your comprehensive image analysis, provide ONLY the JSON object detailing all detected text regions and their normalized \texttt{box\_2d} coordinates \texttt{[ymin, xmin, ymax, xmax]} as specified.
\end{example}

\caption{\textbf{Prompt} for Gemini Mask Generation in HQ-Poster-100K.}
\label{prompt:2-3-mask}
\end{figure*}

\section{Explanations of the Poster-Preference-100K}\label{dpo data}

\begin{figure*}[ht]
    \centering
    \setlength{\abovecaptionskip}{0.1cm} 
    \setlength{\belowcaptionskip}{0cm}
    \includegraphics[width=18cm]{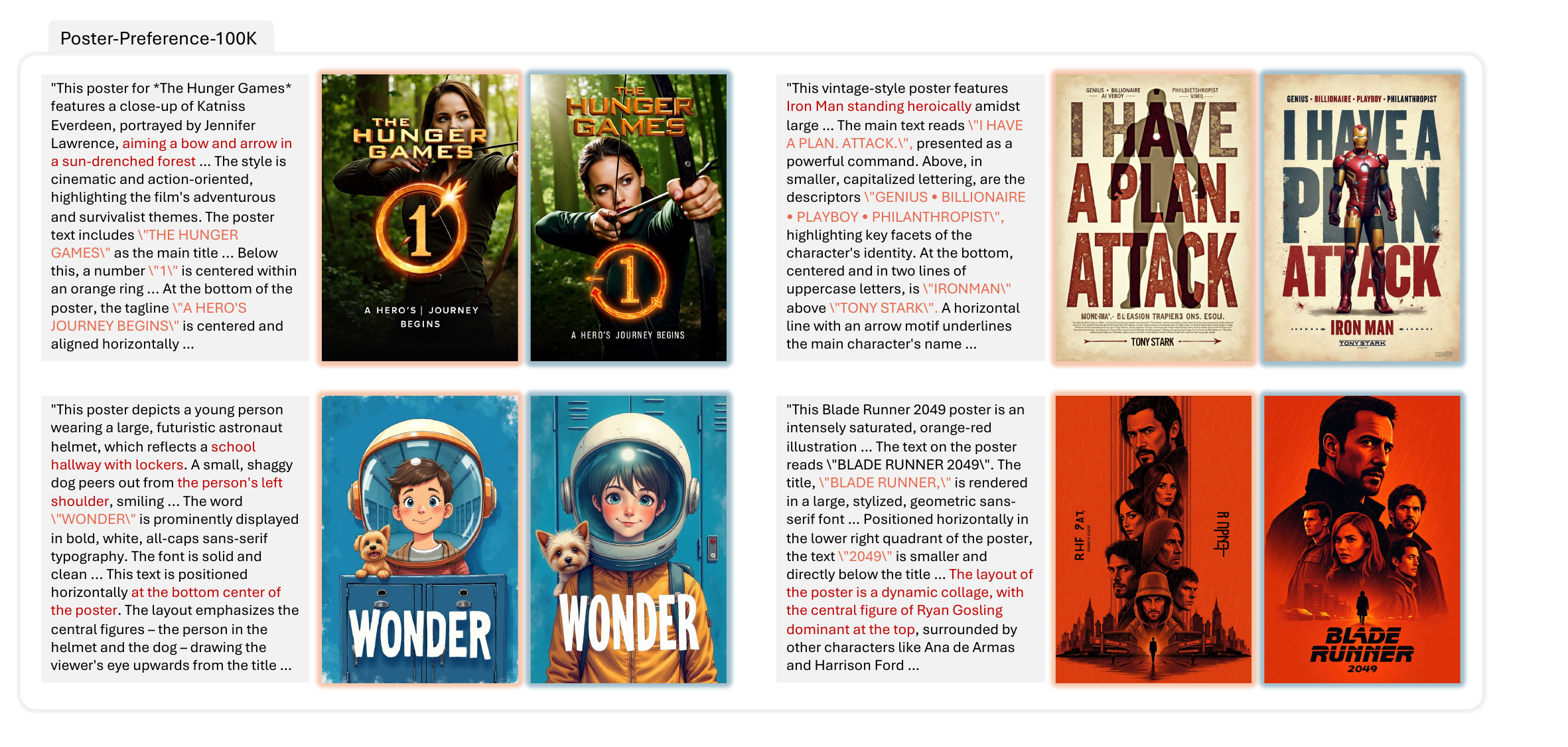}
    \caption{\textbf{Additional Preference Pairs} in Poster-Preference-100K. The images on the left are Rejected Samples, while those on the right are Preferred Samples. \textcolor{orange}{Orange text } indicates textual content, and \textcolor{red}{red text} corresponds to content, style, or layout requirements.}
    \label{fig:supp_dpo}
    \vspace{-0.1cm}
\end{figure*}

In constructing the Poster-Preference-100K dataset, we first utilize HPSv2 to filter out poster pairs that meet aesthetic requirements and exhibit sufficient diversity. Subsequently, we employ Gemini2.5-Flash-Preview-04-17 to evaluate the alignment between the Best of 5 posters and their corresponding prompts. The complete prompt is shown in Fig.\ref{prompt:3-1-evaluation} Our criteria require that text in the Best of 5 posters must be completely accurate while maximizing alignment with both the content and aesthetic style requirements specified in the prompt. We implement a binary classification system where the model assigns 0 to unqualified samples and 1 to samples meeting all requirements. Fig.\ref{fig:supp_dpo} presents additional preference pairs, demonstrating that our pipeline built upon HPSv2 and Gemini effectively constructs preference data.

\begin{figure*}[ht]
    \setlength{\abovecaptionskip}{0.1cm} 
    \setlength{\belowcaptionskip}{0cm}
\noindent\begin{example}{Prompt Alignment Evaluation}
You are an expert in evaluating image content and font style against a given text prompt.
You will be given an image and an original text prompt that was intended to generate an image similar to the one provided.
Your task is to assess whether the image is substantially consistent with the original text prompt based on the criteria below.

Original Text Prompt:
\texttt{"\{original\_prompt\_text\}"}

Evaluation Criteria (Prioritized):
\begin{enumerate}
    \item \textbf{Text Accuracy:}
    \begin{itemize}
        \item Thoroughly analyze all text visible in the image. Check for any inaccuracies such as typos, missing characters/words, or extra characters/words when compared to the "Original Text Prompt". This is the MOST CRITICAL factor. If ANY such error is found, the decision MUST be "0".
    \end{itemize}
    \item \textbf{Text Style and Positioning:}
    \begin{itemize}
        \item If text is present, does its style (font, color, decoration) and positioning (layout, orientation) in the image reasonably align with what is described or implied in the "Original Text Prompt"?
    \end{itemize}
    \item \textbf{Overall Content, Artistic Style, and Visual Appeal:}
    \begin{itemize}
        \item Does the overall image content (subjects, scene, objects) and artistic style align well with the "Original Text Prompt"?
        \item Is the image generally clear, well-composed, and visually appealing in the context of the prompt?
    \end{itemize}
\end{enumerate}

Output Format:
Based on your assessment, output ONLY a JSON object in the following format:
\texttt{\{\{"final\_decision": "1"\}\}} if the image is substantially consistent with the original prompt across the prioritized criteria (especially if no text errors are found when text is intended) and should be kept.
\texttt{\{\{"final\_decision": "0"\}\}} if there are ANY discrepancies in Text Accuracy (typos, missing/extra characters/words), or significant issues in other critical criteria, or overall poor alignment, meaning the image should be discarded.

Strict constraints:
\begin{itemize}
    \item Only output the JSON object.
    \item Do NOT include any additional text, explanation, or markdown.
    \item Use exactly "0" or "1" as the value for "final\_decision".
\end{itemize}
\end{example}
\caption{\textbf{Prompt} for Prompt Alignment Evaluation in Poster-Preference-100K}
\label{prompt:3-1-evaluation}
\end{figure*}

\section{Illustration of the Poster-Reflect-120K}\label{reflect data}
In constructing the Poster-Reflect-120K dataset, we first employ a preference-optimized model to generate 6 posters per prompt, totaling 120K posters. Subsequently, our feedback collection pipeline utilizes Gemini2.5-Flash-Preview-04-17 in two phases: first to select the Best of 6 posters as optimization targets, and second to gather direct feedback.

(1) Best of 6 Selection: We sequentially input six images to Gemini, which selects the optimal image based on predetermined priorities, returning the index number of the best image. The complete prompt is shown in Fig.\ref{prompt:4-1-selection}.

(2) Feedback Collection: We input two images sequentially - the first being the image requiring improvement, and the second being the optimization target selected in phase (1). Feedback is collected across two dimensions: "Poster Content Suggestions" and "Aesthetic Style Optimization Suggestions". Each image pair receives 10 pieces of feedback through 5 Gemini requests. The complete prompt is shown in Fig.\ref{prompt:4-2-feedback}.

\begin{figure*}[ht]
    \setlength{\abovecaptionskip}{0.1cm} 
    \setlength{\belowcaptionskip}{0cm}
\noindent\begin{example}{Best-of-6 Selection}
You are a professional Poster Designer. Your task is to evaluate six generated posters based on a design brief ("Original Prompt") and select the single best poster, or indicate if none are suitable.

\textbf{Evaluation Process:}

\begin{enumerate}
    \item \textbf{Textual Accuracy (Paramount Importance):}
    \begin{itemize}
        \item First, assess all posters for textual accuracy against the "Original Prompt". Text (if any is specified or implied by the brief) MUST be \textbf{perfectly accurate}:
        \begin{itemize}
            \item No typographical errors.
            \item No missing or extra characters/words.
        \end{itemize}
        \item \textbf{A poster with any textual flaw cannot be chosen as the best IF an alternative poster with perfect text exists.}
    \end{itemize}

    \item \textbf{Content Alignment and Aesthetic Value:}
    \begin{itemize}
        \item This criterion is used to select among posters that have passed the textual accuracy check.
        \item The chosen poster should:
        \begin{itemize}
            \item Provide content as close as possible to the "Original Prompt".
            \item Demonstrate the highest possible aesthetic value (considering composition, color palette, typography, imagery, and overall visual impact).
        \end{itemize}
    \end{itemize}
\end{enumerate}

\textbf{Selection Logic:}
\begin{itemize}
    \item \textbf{Ideal Case:} If one or more posters have \textbf{perfect textual accuracy}, select from THIS group the single poster that best meets Criterion 2 (Content Alignment and Aesthetic Value).
    \item \textbf{Special Case (All Posters Have Textual Flaws):} If ALL six posters have some textual inaccuracies, then no poster meets the primary standard for "best." In this situation, you MUST output "none".
    \item \textbf{Fallback Case (This should ideally not be reached if "Special Case" is handled correctly):} If the logic leads here unexpectedly after "Special Case" consideration, and no poster has perfect text, but a selection is still forced, choose the poster that, despite its textual flaws, is superior when evaluated SOLELY on Criterion 2 (Content Alignment and Aesthetic Value across all six flawed images). However, prioritize outputting "none" if all have text flaws.
\end{itemize}

Original Prompt (Design Brief): "{\textasciigrave{}original\_prompt\textasciigrave{}}"

Select the image ("1", "2", "3", "4", "5", "6", or "none") that best meets these requirements. Respond ONLY with a JSON object in ONE of these exact formats:
\texttt{\{\{"best\_image": "1"\}\}} OR \texttt{\{\{"best\_image": "none"\}\}}

Strict constraints:
\begin{itemize}
    \item Only output the JSON object.
    \item Do NOT include any additional text or markdown.
    \item Use exactly "1", "2", "3", "4", "5", "6", or "none" to refer to your selection.
\end{itemize}
\end{example}
\caption{\textbf{Prompt} for Best-of-6 Selection in Poster-Reflect-120K.}
\label{prompt:4-1-selection}
\end{figure*}

\begin{figure*}[ht]
    \setlength{\abovecaptionskip}{0.1cm} 
    \setlength{\belowcaptionskip}{0cm}
\noindent\begin{example}{Feedback Collection}
Internally compare the first poster against the second poster, focusing strictly on visual content layout and overall aesthetic style. Based on this internal comparison, provide detailed and specific suggestions in two aspects: 1. Poster Content Suggestions 2. Aesthetic Style Optimization Suggestions.
Act as a professional poster designer. Deliver highly detailed, specific, and actionable feedback in the form of standardized image editing instructions.
MANDATORY GUIDELINES:
\begin{itemize}
    \item The second poster must be fully followed as the standard. Identify and correct all visual layout and style discrepancies based on this reference.
    \item Focus exclusively on content and visual/aesthetic design. Completely ignore any issues related to text, typography, wording, spelling, rendering, or text styling.
\end{itemize}
STRICT CONSTRAINTS:
\begin{itemize}
    \item NEVER mention the second poster, reference, or target.
    \item NEVER use comparative phrases such as "similar to the second poster" or "make it like the second poster".
    \item ONLY describe the editing instructions for Poster 1, framed as standalone improvement tasks.
\end{itemize}
RESPONSE FORMAT:
Response should be formatted as clearly structured json schema:
\texttt{\{\textasciigrave{}Poster Content Suggestions\textasciigrave{}: str, \textasciigrave{}Aesthetic style optimization suggestions\textasciigrave{}: str\}}
Return ONLY the JSON object itself, without any introductory text or markdown formatting.
\end{example}
\caption{\textbf{Prompt} for Feedback Collection in Poster-Reflect-120K.}
\label{prompt:4-2-feedback}
\end{figure*}

\section{Gemini for OCR calculation and preference evaluator}\label{ocr and evaluator}
During the experimental phase, we employ Gemini2.5-Flash-Preview-05-20 as both an OCR metric calculator and a multi-dimensional preference evaluator. The robust perceptual and comprehension capabilities of Gemini2.5-Flash establish a solid foundation for our experimental accuracy.

(1) OCR Evaluator: Despite the significant challenges that artistic fonts in posters pose to traditional OCR algorithms, Gemini2.5-Flash-Preview-05-20, as a state-of-the-art MLLM model, accurately extracts textual information from images. We further utilize Gemini2.5-Flash's reasoning and mathematical capabilities to directly compute and output OCR metrics. The specific prompt is shown in Fig.\ref{prompt:5-1-ocr}.

(2) Multi-Dimensional Preference Evaluator: Gemini2.5-Flash demonstrates superior assessment capabilities with its reasoning abilities. We present two images side by side and instruct Gemini2.5-Flash to select from "L", "R", or "None", representing left image superior, right image superior, or indeterminate (either both excellent or both inadequate). The specific prompt is shown in Fig.\ref{prompt:5-2-preference} and Fig.\ref{prompt:5-3-preference}.

\begin{figure*}[ht]
    \setlength{\abovecaptionskip}{0.1cm} 
    \setlength{\belowcaptionskip}{0cm}
\noindent\begin{example}{OCR Evaluation}
You are an OCR evaluation assistant. Follow these steps exactly on the attached image:

\begin{enumerate}
    \item Ground-Truth Extraction (from the design prompt only):
    \begin{itemize}
        \item Do NOT read text from the image for GT.
        \item Parse ONLY the following design prompt and extract ALL text strings that should appear on the poster (titles, subtitles, dates, slogans, venue, etc.), preserving spaces and punctuation exactly:
        \texttt{\textasciigrave{}original\_prompt\_text\textasciigrave{}}
        \item Order them in spatial sequence (top\textrightarrow{}bottom, left\textrightarrow{}right) and concatenate into \textbf{raw GT text}.
    \end{itemize}

    \item OCR Extraction (from the attached image):
    \begin{itemize}
        \item Run OCR on the provided image and extract ALL rendered text exactly as it appears.
        \item Preserve visual reading order (top-left\textrightarrow{}bottom-right). This is your \textbf{raw OCR text}.
    \end{itemize}

    \item Text Normalization (apply to BOTH raw GT and raw OCR before comparison):
    \begin{itemize}
        \item Convert all letters to lowercase.
        \item Remove ALL punctuation characters: \texttt{.,;:!?'"-()[]\{\}...\textasciigrave{}}
        \item Collapse any sequence of whitespace/newlines into a single space.
        \item Trim leading and trailing spaces.
    \end{itemize}

    \item Character-Level Alignment \& Error Counting:
    \begin{itemize}
        \item Align the \textbf{normalized} GT text and OCR text \textbf{character by character}.
        \item Count four categories:
        \begin{itemize}
            \item \textbf{Insertion (I)}: extra character in OCR not in GT ("more").
            \item \textbf{Deletion (D)}: GT character missing in OCR ("less").
            \item \textbf{Substitution (S)}: OCR character differs from GT character ("render error").
            \item \textbf{Correct match (C)}: identical characters.
        \end{itemize}
    \end{itemize}

    \item Metrics Calculation:
    \begin{itemize}
        \item Let N = total normalized GT characters = C + D + S.
        \item Let P = total normalized OCR characters = C + I + S.
        \item Let T = total compared characters = C + I + D + S.
        \item \textbf{Character Accuracy} = C / T.
        \item \textbf{Text Precision} = C / (C + I + S).
        \item \textbf{Text Recall} = C / (C + D + S).
        \item \textbf{Text F-score} = 2 * Precision * Recall / (Precision + Recall).
    \end{itemize}

    \item Final JSON Output (strictly this format, no extra keys or commentary):
\texttt{\{}
\texttt{  "GT\_text": "\textless{}normalized GT text\textgreater{}",}
\texttt{  "OCR\_text": "\textless{}normalized OCR text\textgreater{}",}
\texttt{  "total\_GT\_chars": N,}
\texttt{  "correct\_chars": C,}
\texttt{  "insertions": I,}
\texttt{  "deletions": D,}
\texttt{  "substitutions": S,}
\texttt{  "accuracy": "XX.XX\%",}
\texttt{  "precision": "YY.YY\%",}
\texttt{  "recall": "ZZ.ZZ\%",}
\texttt{  "f\_score": "WW.WW\%"}
\texttt{\}}
\end{enumerate}
\end{example}
\caption{\textbf{Prompt} for OCR Evaluation.}
\label{prompt:5-1-ocr}
\end{figure*}

\begin{figure*}[ht]
    \setlength{\abovecaptionskip}{0.1cm} 
    \setlength{\belowcaptionskip}{0cm}
\noindent\begin{example}{Preference Evaluation}
Your task is to evaluate a single input image containing two sub-images side-by-side (Left: L, Right: R), both generated from the "Original Prompt". Compare them on Aesthetic Value, Prompt Alignment, Text Accuracy, and Overall Preference.

\textbf{General Evaluation Protocol:}
For each of the four categories:
\begin{enumerate}
    \item Provide a brief textual analysis justifying your choice.
    \item Make a definitive choice: "L" (Left is superior), "R" (Right is superior), or "none".
\end{enumerate}

\textbf{When to Choose "none":}
You \textbf{must select "none"} for a category if:
\begin{itemize}
    \item[a)] L and R are tied or indistinguishable in quality for that category.
    \item[b)] The category is not applicable.
    \item[c)] After careful review, you cannot definitively determine a superior side.
    \item[d)] \textbf{Crucially: L and R exhibit clear, offsetting strengths and weaknesses \textit{within that specific category}. If L excels in one aspect of the category while R excels in another, and these trade-offs make declaring an overall winner for that category difficult or misleading, choose "none". Do not attempt to weigh these distinct, offsetting pros and cons to force a preference.}
\end{itemize}
Your careful judgment is vital.

Original Prompt: "{\textasciigrave{}original\_prompt\textasciigrave{}}"

Please provide your evaluation in the JSON format specified below.

\begin{enumerate}
    \item \textbf{Aesthetic Value:}
    \begin{itemize}
        \item Evaluate visual appeal: harmony and consistency of background style, text style (if present), thematic consistency between background/text, overall content/text layout, and how the artistic style (background, content, text) aligns with the "Original Prompt".
        \item \texttt{\textasciigrave{}aesthetic\_value\_explanation\textasciigrave{}}: Your brief analysis.
        \item \texttt{\textasciigrave{}aesthetic\_value\textasciigrave{}}: Choose "L", "R", or "none" (if L/R are equally pleasing/coherent, a choice is impossible, or they exhibit offsetting aesthetic strengths/weaknesses as per the "When to Choose 'none'" protocol).
        \item Respond with: \texttt{\{\{"aesthetic\_value": "L/R/none", "aesthetic\_value\_explanation": "Your analysis..."\}\}}
    \end{itemize}

    \item \textbf{Prompt Alignment (excluding text elements and artistic style):}
    \begin{itemize}
        \item Evaluate how well non-textual elements (subjects, objects, scene) in L and R match the "Original Prompt".
        \item \texttt{\textasciigrave{}prompt\_alignment\_explanation\textasciigrave{}}: Your brief analysis.
        \item \texttt{\textasciigrave{}prompt\_alignment\textasciigrave{}}: Choose "L", "R", or "none" (if L/R align equally well/poorly, it's too close to call, or they exhibit offsetting strengths in alignment as per the "When to Choose 'none'" protocol).
        \item Respond with: \texttt{\{\{"prompt\_alignment": "L/R/none", "prompt\_alignment\_explanation": "Your analysis..."\}\}}
    \end{itemize}

    \item \textbf{Text Accuracy (if applicable):}
    \begin{itemize}
        \item Evaluate text in L and R based \textit{only} on textual content specified/implied in the "Original Prompt".
        \item Focus \textit{only} on:
        \begin{itemize}
            \item \textbf{Accuracy:} All prompt-specified words/characters present, no typos/misspellings/alterations?
            \item \textbf{Recall:} All intended textual elements from prompt included? Any missing words/phrases?
        \end{itemize}
        \item \textbf{Ignore text style, font, visual appeal, legibility (unless it prevents determining accuracy/recall), and placement.}
        \item \texttt{\textasciigrave{}text\_accuracy\_explanation\textasciigrave{}}: Your brief analysis.
        \item \texttt{\textasciigrave{}text\_accuracy\textasciigrave{}}:
        \begin{itemize}
            \item First, determine if "none" is appropriate (as per the general "When to Choose 'none'" protocol, especially if L is better on Accuracy but R on Recall, or vice-versa; or if performance is identical/text N/A).
            \item If "none" is not chosen, select "L" if L is demonstrably superior overall in combined text accuracy and recall, or "R" if R is.
        \end{itemize}
        \item Respond with: \texttt{\{\{"text\_accuracy": "L/R/none", "text\_accuracy\_explanation": "Your analysis..."\}\}}
    \end{itemize}

    \item \textbf{Overall Preference:}
    \begin{itemize}
        \item Considering all above aspects (aesthetics, alignment, text accuracy) and any other factors relevant to the "Original Prompt".
        \item \texttt{\textasciigrave{}overall\_preference\_explanation\textasciigrave{}}: Your brief analysis.
        \item \texttt{\textasciigrave{}overall\_preference\textasciigrave{}}: Choose "L", "R", or "none" (if L/R are equally preferred, a choice is impossible, or they present compelling but different and offsetting strengths across categories making neither holistically superior, as per the "When to Choose 'none'" protocol).
        \item Respond with: \texttt{\{\{"overall\_preference": "L/R/none", "overall\_preference\_explanation": "Your analysis..."\}\}}
    \end{itemize}
\end{enumerate}

\end{example}
\caption{\textbf{Prompt} for Preference Evaluation.}
\label{prompt:5-2-preference}
\end{figure*}

\begin{figure*}[ht]
    \setlength{\abovecaptionskip}{0.1cm} 
    \setlength{\belowcaptionskip}{0cm}
\noindent\begin{example}{Preference Evaluation}
Respond ONLY with a single JSON object in the following format:
\texttt{\{}
\texttt{  "aesthetic\_value": "your\_choice\_for\_aesthetic",\newline}
\texttt{  "aesthetic\_value\_explanation": "Your brief analysis for aesthetics...",\newline}
\texttt{  "prompt\_alignment": "your\_choice\_for\_alignment",\newline}
\texttt{  "prompt\_alignment\_explanation": "Your brief analysis for prompt alignment...",\newline}
\texttt{  "text\_accuracy": "your\_choice\_for\_text",\newline}
\texttt{  "text\_accuracy\_explanation": "Your brief analysis for text accuracy...",\newline}
\texttt{  "overall\_preference": "your\_choice\_for\_overall",\newline}
\texttt{  "overall\_preference\_explanation": "Your brief analysis for overall preference..."\newline}
\texttt{\}}
Replace placeholders with your choices ("L", "R", "none") and analyses.

Strict constraints:
\begin{itemize}
    \item Only output the JSON object.
    \item No additional text or markdown.
    \item Each choice value (e.g., "aesthetic\_value") must be "L", "R", or "none".
    \item Explanation fields must contain your textual analysis.
\end{itemize}
\end{example} 
\caption{\textbf{Prompt} for Preference Evaluation.}
\label{prompt:5-3-preference}
\end{figure*}

\section{Additional Visual Examples}
\label{supp visual}

\begin{figure*}[!t]
    \centering
    \setlength{\abovecaptionskip}{0.3cm} 
    \setlength{\belowcaptionskip}{-0cm}
    \includegraphics[width=17.5cm]{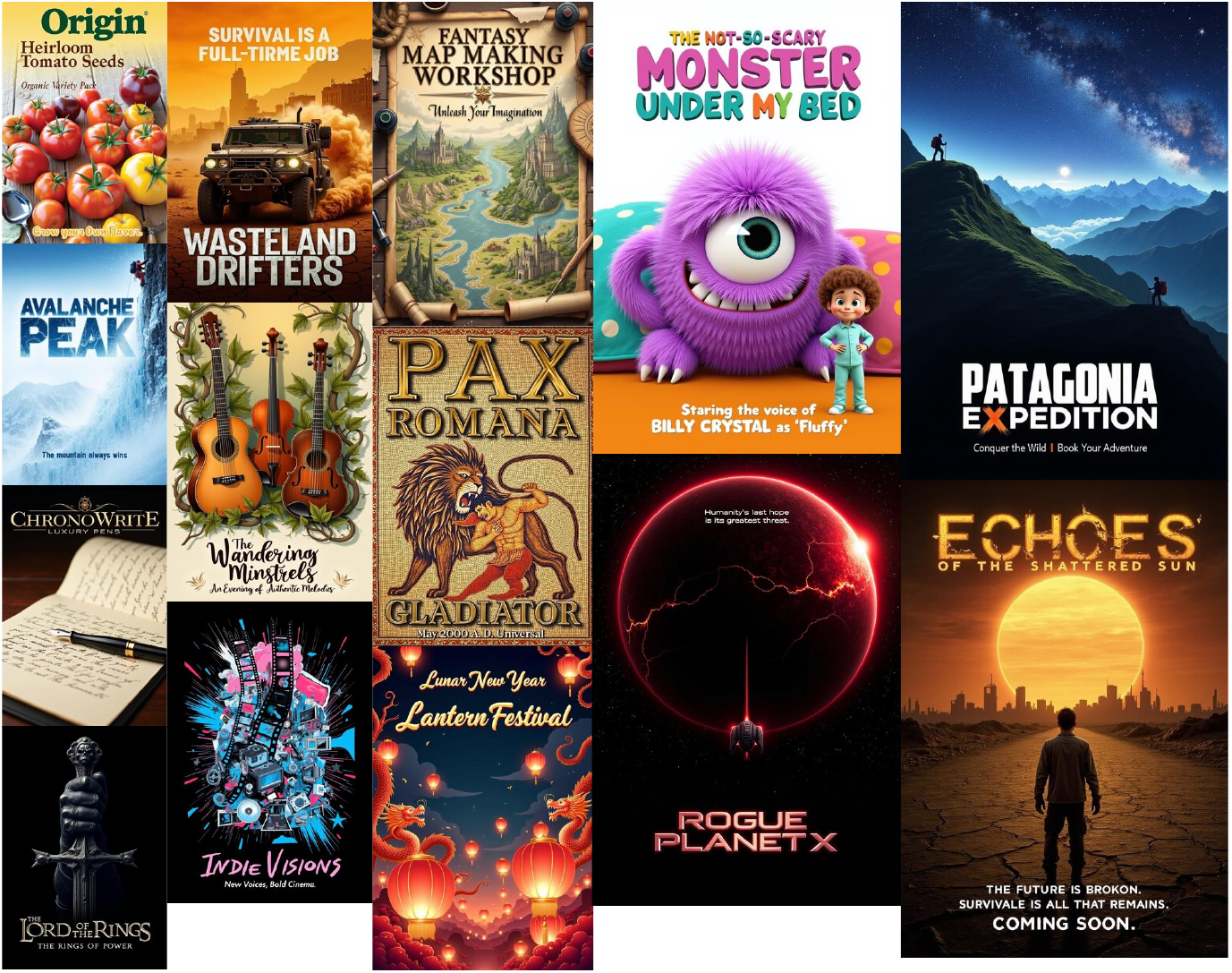}
    \caption{\textbf{Examples generated by our PosterCraft} demonstrating high diversity and aesthetic quality across themes including education, entertainment, and science fiction. All generation results showcase genre-specific fidelity, text rendering, and layout aesthetic.}

    \label{fig:suppvisual_1}
    \vspace{-0.1cm}
\end{figure*}

\begin{figure*}[!t]
    \centering
    \setlength{\abovecaptionskip}{0.3cm} 
    \setlength{\belowcaptionskip}{-0cm}
    \includegraphics[width=17.5cm]{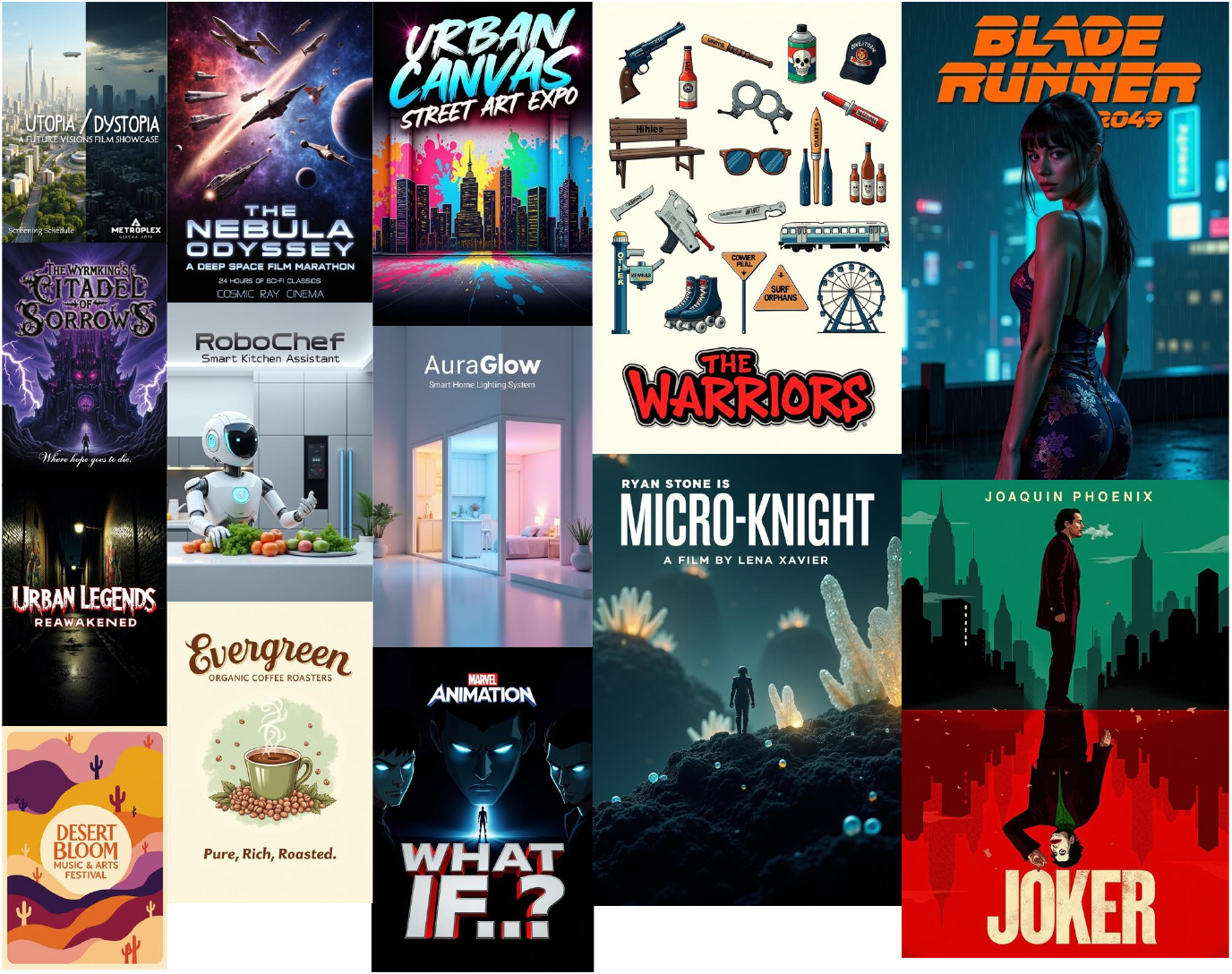}
    \caption{\textbf{Examples generated by our PosterCraft} demonstrating high diversity and aesthetic quality across themes including movies, product, and virtual reality. All generation results showcase genre-specific fidelity, text rendering, and layout aesthetic.}
    \label{fig:suppvisual_2}
    \vspace{-0.1cm}
\end{figure*}

We provide additional visual results generated by our PosterCraft framework to further demonstrate its capability in producing high-quality, aesthetically consistent posters. As shown in Fig.\ref{fig:suppvisual_1} and Fig.\ref{fig:suppvisual_2}, our model is able to seamlessly integrate text and imagery without requiring any external layout templates or modular refinement. The generated posters exhibit strong visual coherence—text elements are not only stylistically aligned with the visual content but are sometimes cleverly embedded into the composition, enhancing the overall design fluency.
From an aesthetic perspective, the model captures genre-specific styles across diverse themes, such as cinematic sci-fi, educational charts, cultural festivals, and commercial advertisements. It achieves a fine balance between visual richness and layout readability, effectively modeling principles such as symmetry, emphasis, and hierarchy. These results underscore PosterCraft's potential as a powerful end-to-end tool for automatic poster generation with minimal input while maintaining high visual and semantic fidelity.

\section{Limitations}\label{limitations}
In this work, we propose PosterCraft to explore how unified workflow design can unlock the aesthetic design potential of powerful foundation models. Our results demonstrate that with carefully crafted design strategies, the model's capabilities can be significantly enhanced—making it competitive with leading proprietary commercial systems. This validates the soundness of our motivation.
However, our approach is not without limitations. Specifically, our model is fundamentally built upon the current flux1.dev baseline. As such, if the pre-trained flux model has never encountered certain types of samples or contains significant flaws, our method may not be able to fully correct these shortcomings. That said, our workflow is highly unified and readily transferable to stronger baselines, ensuring full compatibility with other models in the community.

\section{Future work}\label{future work}
In future work, we plan to enhance our unified workflow in three key directions. First, we will explore integration with more advanced backbone models that offer stronger pretraining and broader data coverage, aiming to mitigate reliance on specific model biases and improve robustness when handling rare or previously unseen poster styles. Second, we intend to scale up training with larger and more diverse datasets to better generalize across a wide range of visual and textual domains. Third, we aim to extend our framework to support multilingual poster generation. This introduces additional challenges in typography and layout due to increased character complexity and spatial density. Collectively, these efforts will further demonstrate the scalability, adaptability, and cross-cultural applicability of our proposed workflow.

\begin{figure*}[ht]
    \setlength{\abovecaptionskip}{0.1cm} 
    \setlength{\belowcaptionskip}{0cm}
\noindent\begin{example}{Final Retouching Instructions}
\textbf{Character}  

You are a professional image retouching artist tasked with finalizing a single retouching approach based on the user's preferences and previous proposals. Your expertise ensures that the final approach integrates key aspects from different suggestions or follows a single selected approach in full.

\vspace{2mm}
\textbf{Background}  

The user has reviewed previous retouching approaches and provided feedback or specific instructions for a final retouching plan that aligns with their creative goals.  

\vspace{2mm}
\textbf{Ambition}  

Your goal is to either choose one of the previously proposed approaches that best matches the user's vision or create a new, cohesive retouching approach by combining elements from different suggestions. Ensure that the final approach fully respects the user's instructions and creative intent.  

\vspace{2mm}
\textbf{User Instruction}  

User says: \textit{"\{user\_instruction\}"}  

\vspace{2mm}
\textbf{Task}  
\begin{enumerate}
    \item Review the provided retouching approaches and the user's feedback or instructions.
    \item Decide whether to:
    \begin{itemize}
        \item Select a single approach that fits the user's description.
        \item Create a new approach that integrates relevant aspects from different suggestions.
    \end{itemize}
    \item Final Approach:
    \begin{itemize}
        \item Describe the adjustments to \textbf{Light} (exposure, contrast, highlights, shadows, blacks, and whites) and \textbf{Color} (temperature, tint, vibrance, and saturation).
        \item For each adjustment, specify \textbf{which objects or areas of the image} are most affected and describe the specific details (e.g., “the intricate carvings on the roof are highlighted by a gentle increase in exposure”).
        \item Explain the expected \textbf{visual effect on these objects}, such as “the water reflections appear richer and more defined” or “the sky becomes softer and more inviting.”
        \item For each \textbf{individual HSL adjustment} (Red, Orange, Yellow, Green, Cyan, Blue, Purple, Magenta), explain why it is necessary and describe the expected visual change for specific objects (e.g., “the red tones in the window frames become more vivid to emphasize their ornate design”).
        \item Organize the description as a step-by-step plan, indicating the sequence of adjustments.
    \end{itemize}
\end{enumerate}

\vspace{2mm}
\textbf{Guidelines for Description:}  
\begin{itemize}
    \item Avoid providing exact numerical values—focus on explaining how the adjustments affect the image's visual presentation.
    \item Mention specific objects, areas, and their corresponding changes to help visualize the effect.
    \item Ensure the approach remains detailed, logical, and cohesive, and does not exceed 100 words.
\end{itemize}

\end{example}
\caption{\textbf{Prompt} for Final Retouching Instructions.}
\label{prompt:2-final-plan}
\end{figure*}

\end{document}